\documentclass[10pt,journal,compsoc]{IEEEtran}
\usepackage{url}            
\usepackage{booktabs}       
\usepackage{amsfonts}       
\usepackage{nicefrac}       
\usepackage{microtype}      
\usepackage{xcolor}         
\usepackage{amsmath}
\usepackage{graphicx}
\usepackage{algorithm}
\usepackage{algorithmic}
\usepackage{multicol}
\usepackage{multirow}
\usepackage{stfloats}

\ifCLASSOPTIONcompsoc
  \usepackage[nocompress]{cite}
\else
  \usepackage{cite}
\fi
\ifCLASSINFOpdf
\else
\fi
\usepackage[colorlinks,linkcolor=red]{hyperref}       
\usepackage{url}   

\begin{document}
%
\title{Unsupervised 3D Pose Transfer with Cross Consistency and Dual Reconstruction}
%
%
%
%

\author{Chaoyue Song, Jiacheng Wei, Ruibo Li, Fayao Liu and Guosheng Lin
\IEEEcompsocitemizethanks{\IEEEcompsocthanksitem Chaoyue Song (Email: chaoyue002@e.ntu.edu.sg), Ruibo Li and Guosheng Lin are with S-Lab, Nanyang Technological University, Singapore and the School of Computer Science and Engineering, Nanyang Technological University, Singapore.
\IEEEcompsocthanksitem Jiacheng Wei is with the School of Electrical and Electronic Engineering, Nanyang Technological University, Singapore.
\IEEEcompsocthanksitem Fayao Liu is with the Institute for Inforcomm Research, A*STAR, Singapore.
\IEEEcompsocthanksitem Corresponding author: Guosheng Lin (Email: gslin@ntu.edu.sg)}}
%
%

\markboth{Journal of \LaTeX\ Class Files,~Vol.~14, No.~8, August~2015}%
{Unsupervised 3D Pose Transfer with Cross Consistency and Dual Reconstruction}
%



\IEEEtitleabstractindextext{%
\begin{abstract}
The goal of 3D pose transfer is to transfer the pose from the source mesh to the target mesh while preserving the identity information (e.g., face, body shape) of the target mesh. Deep learning-based methods improved the efficiency and performance of 3D pose transfer. However, most of them are trained under the supervision of the ground truth, whose availability is limited in real-world scenarios. In this work, we present \textit{X-DualNet}, a simple yet effective approach that enables unsupervised 3D pose transfer. In \textit{X-DualNet}, we introduce a generator $G$ which contains correspondence learning and pose transfer modules to achieve 3D pose transfer. We learn the shape correspondence by solving an optimal transport problem without any key point annotations and generate high-quality meshes with our elastic instance normalization (ElaIN) in the pose transfer module. With $G$ as the basic component, we propose a cross consistency learning scheme and a dual reconstruction objective to learn the pose transfer without supervision. Besides that, we also adopt an as-rigid-as-possible deformer in the training process to fine-tune the body shape of the generated results. Extensive experiments on human and animal data demonstrate that our framework can successfully achieve comparable performance as the state-of-the-art supervised approaches. Project page: \url{https://chaoyuesong.github.io/X-DualNet}.
\end{abstract}

}

\maketitle

\IEEEdisplaynontitleabstractindextext

%
\IEEEpeerreviewmaketitle

\IEEEraisesectionheading{\section{Introduction}\label{sec:intro}}

%
%
%
%

\begin{figure*}
  \centering
  \includegraphics[scale=0.3]{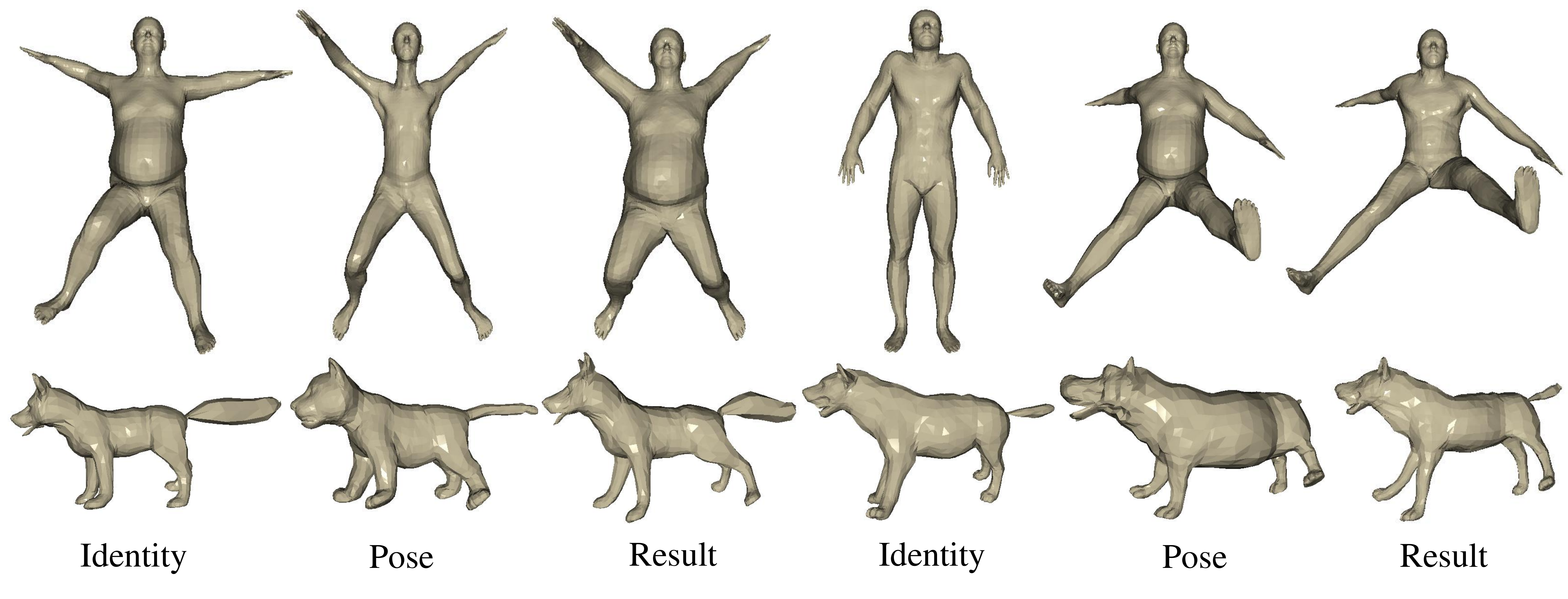}
  \caption{With a target mesh and a source mesh, \textit{X-DualNet} provides an unsupervised solution to transfer the pose from the source to the target and preserve the identity information of the target mesh. In the first row, the human meshes are from SMPL \cite{loper2015smpl} and FAUST \cite{bogo2014faust}. In the second row, the animal meshes are from SMAL \cite{zuffi20173d}. X-DualNet can achieve the 3D pose transfer successfully on different human and animal datasets.}
  \label{intro}
\end{figure*}

\IEEEPARstart{3}{D} Pose Transfer is a crucial task in vision and computer graphics communities which has many applications in augmented reality, robotics, and video games. As shown in Figure \ref{intro}, given a source mesh that provides the pose and a target mesh that provides the identity (e.g., face, body shape), 3D pose transfer aims to transfer the pose from the source to the target and preserve the identity information of the target mesh.

Traditional method \cite{sumner2004deformation} defines an optimization problem to address 3D pose transfer. They apply the affine transformations computed from the deformation of the source mesh to the target one, which are time-consuming and always need key points labeling to build the shape correspondence. With the huge success of neural networks, many methods \cite{wang2020neural, song20213d} were proposed to achieve 3D pose transfer in a deep learning manner. Wang \textit{et al.} \cite{wang2020neural} re-purpose the latest style transfer methods for 3D pose transfer. The performance of their method is always unsatisfactory since they do not consider any correspondence between different human or animal meshes. Besides that, they also need the supervision of the ground truth label, which is hard to acquire in real-world scenarios. 


In this work, we present \textit{X-DualNet}, an effective approach to achieve unsupervised 3D pose transfer. With a generator $G$ which contains correspondence learning and pose transfer modules as the basic component, we propose a cross consistency learning scheme and a dual reconstruction objective to learn the pose transfer without supervision. Specifically, $G$ takes the vertex coordinates of the input meshes and extracts their features. Then we solve an optimal transport problem to build the shape correspondence in the correspondence learning module and achieve the 3D pose transfer using the proposed elastic instance normalization (ElaIN) in the pose transfer module. Similar to \cite{wang2020neural, song20213d}, our approach can handle the training and test meshes with different numbers of vertices or triangles.

For the correspondence learning module, we treat shape correspondence learning as an optimal transport problem to learn the correspondence between meshes. Our network takes vertex coordinates of identity and pose meshes as inputs. We extract deep features at each vertex using point cloud convolutions and compute a matching cost between the vertex sets with the extracted features. Our goal is to minimize the matching cost to get an optimal matching matrix. With the optimal matching matrix, we warp the pose mesh and obtain the warped mesh which is actually a redistribution of the vertices of the pose mesh. We then transfer the pose from the warped mesh to the identity mesh with a set of elastic instance normalization residual blocks. The modulation parameters in the normalization layers are learned with our proposed \textit{Elastic Instance Normalization (ElaIN)}. In order to generate smoother meshes with more details, we introduce a channel-wise weight in ElaIN to adaptively blend feature statistics of original features and the learned parameters from external data, which helps to keep the consistency and continuity of the original features.

In \textit{X-DualNet}, with a main generator $G_{AB}$, we take two meshes $M_{A}$ and $M_{B}$ as inputs and generate the output $M_{output} = G_{AB}(M_{A}, M_{B})$. Here, we view $M_{A}$ as the identity mesh and $M_{B}$ as the pose mesh respectively. Besides that, we also add an ARAP deformer in the training process to fine-tune the body shape of the generated results and get the deformed output $\hat{M}_{output}$. Then we adopt an auxiliary generator $G^{\prime}_{A}$ to reconstruct the mesh $M_{A}$. Specifically, $G^{\prime}_{A}$ takes $M_{A}$ as the pose mesh and $\hat{M}_{output}$ as the identity mesh respectively to generate $M^{\prime}_{A} = G^{\prime}_{A}(\hat{M}_{output}, M_{A})$. Similarly, we can reconstruct $M_{B}$ with another auxiliary generator $G^{\prime}_{B}$ to generate $M^{\prime}_{B} = G^{\prime}_{B}(M_{B}, \hat{M}_{output})$. Note that $G_{AB}$, $G^{\prime}_{A}$ and $G^{\prime}_{B}$ share the same network parameters. The pose transfer procedure is consistent across the main generator and two auxiliary generators.
To achieve this dual reconstruction objective, we introduce a reconstruction loss that encourages $M^{\prime}_{A} \approx M_{A}$ and $M^{\prime}_{B} \approx M_{B}$. Based on this cross consistency learning scheme and the dual reconstruction objective, our \textit{X-DualNet} can learn the 3D pose transfer without any supervision. 

In summary, our main contributions are:

$\bullet$ We present \textit{X-DualNet}, an unsupervised deep learning framework to solve the 3D pose transfer problem in an end-to-end fashion.

$\bullet$ In \textit{X-DualNet}, we introduce a generator $G$ that contains correspondence learning and pose transfer modules to achieve the 3D pose transfer. We learn the shape correspondence by solving an optimal transport problem without any key point annotations and generate high-quality meshes with our proposed elastic instance normalization in the pose transfer module. To the best of our knowledge, our method is the first to learn the correspondence between different meshes and transfer the poses jointly in the 3D pose transfer task. 

$\bullet$ With the generator $G$ as the basic component, we propose a cross consistency learning scheme and a dual reconstruction objective to learn the pose transfer without supervision. Besides that, we also adopt an as-rigid-as-possible deformer to fine-tune the body shape of the generated results.

$\bullet$ Through extensive experiments on human and animal meshes, we demonstrate \textit{X-DualNet} achieves comparable performance as the state-of-the-art supervised approaches qualitatively and quantitatively and even outperforms some of them. In addition, our framework has better generalization capability than supervised methods.

This work is an extended version of 3D-CoreNet \cite{song20213d}. The code is available at \url{https://github.com/ChaoyueSong/3d-corenet}.

\section{Related work}
\subsection{Deep networks for 3D representations}
Recently, many different methods have been introduced to analyze 3D data. VoxNet \cite{maturana2015voxnet} and 3DShapeNets \cite{wu20153d} were introduced to learn on volumetric grids. However, the extra computation cost makes it difficult for them to apply to complex data. Due to the irregular format of point cloud, PointNet \cite{qi2017pointnet} processes each point independently followed by a symmetry function, but it cannot captures the local structure. To address this problem, SO-Net\cite{li2018so} and PointNet++ \cite{qi2017pointnet++} propose to aggregate local structures with nearest neighbors. For 3D meshes, Meshnet\cite{feng2019meshnet} and mesh VAE\cite{tan2018variational} propose mesh neural networks to capture and learn features of polygon faces. However, their methods require many computing resources because of their fully-connected layers. There are also many works using graph convolutions with mesh sampling operators \cite{garland1997surface} to implement the hierarchical structure, such as CoMA \cite{ranjan2018generating}, SpiralNet \cite{lim2018simple} and SpiralNet++ \cite{gong2019spiralnet++}, etc. Their methods all require the mesh datasets registered to a template for training. In this work, we do not need the mesh template and our approach is able to handle non-registered meshes with different numbers of vertices or triangles.

\subsection{3D pose transfer}
3D pose transfer has been widely studied in vision and graphics communities. Many traditional methods require additional information to build the correspondence between different meshes or generate the final results. DT \cite{sumner2004deformation} and Yang \textit{et al.} \cite{yang2018biharmonic} need users to label the corresponding vertices. Baran \textit{et al.} \cite{baran2009semantic} infer semantic correspondence between the shape spaces of two characters by providing examples of corresponding poses. Chu \textit{et al.} \cite{chu2010example} propose to use some examples that are similar to the sources in poses to generate results, their method has difficulties transferring pose automatically.   
Recently, deep neural networks have shown powerful learning and generalization capability to solve 3D pose transfer problems. VC-GAN \cite{gao2018automatic} follows the idea of CycleGAN \cite{Zhu_2017_ICCV} to help 3D pose transfer, but their method relies heavily on the training data and cannot work on new shapes because of the visual similarity. Wang \textit{et al.} \cite{yifan2020neural} extend a traditional cage-based deformation method but need user labeling to handle the differences between meshes. Wang \textit{et al.} \cite{wang2020neural} re-purpose the latest style transfer techniques for 3D pose transfer, the performance of their model is always not satisfied since they do not consider any correspondence between the identity and the pose meshes. To address this problem, we learn the shape correspondence of different human or animal meshes. The previous solutions either require additional information or ground truth to instruct the training process. In this work, we propose to achieve the 3D pose transfer in an unsupervised way with the cross consistency learning scheme and a dual reconstruction objective. 

\subsection{Correspondence learning}
In CoCosNet \cite{zhang2020cross} and CT-Net \cite{Yang_2021_CVPR}, they introduced a correspondence network based on the correlation matrix between images without any constraints. To learn a better matching, we proposed to use optimal transport to learn the correspondence between meshes. Recently, optimal transport has received great attention in various computer vision tasks. Courty \textit{et al.} \cite{courty2016optimal} perform the alignment of the representations in the source and target domains by learning a transportation plan. Su \textit{et al.} \cite{su2015optimal} compute the optimal transport map to deal with the surface registration and shape space problem. Other applications include generative model \cite{arjovsky2017wasserstein, bunne2019learning, deshpande2019max, wu2019sliced}, scene flow \cite{puy2020flot, Li_2021_CVPR}, point cloud segmentation \cite{shi2022weakly}, semantic correspondence \cite{liu2020semantic} and etc.

\subsection{Conditional normalization layers}
After normalizing the activation value, conditional normalization uses the modulation parameters calculated from the external data to denormalize it. Adaptive Instance Normalization (AdaIN) \cite{huang2017arbitrary} aligns the mean and variance between content and style image which achieves arbitrary style transfer. Soft-AdaIN \cite{chen2020cartoonrenderer} introduces a channel-wise weight to blend feature statistics of content and style images to preserve more details for the results. Spatially-Adaptive Normalization (SPADE) \cite{park2019semantic} can better preserve semantic information by not washing away it when applied to segmentation masks. SPAdaIN \cite{wang2020neural} changes batch normalization \cite{ioffe2015batch} in SPADE to instance normalization \cite{ulyanov2016instance} for 3D pose transfer. However, it will break the consistency and continuity of the feature map when doing the denormalization, which has a bad influence on the mesh smoothness and detail preservation. To address this problem, our ElaIN introduces an adaptive weight to implement the denormalization.  

\subsection{Unsupervised learning for 3D shapes}
CorrNet3D \cite{zeng2021corrnet3d} proposes to learn the shape correspondence without supervision using deformation-like reconstruction. NeuroMorph \cite{eisenberger2021neuromorph} solves the correspondence and interpolation problems with geometric priors for regularization. Zhou \textit{et al.} \cite{zhou2020unsupervised} and Chen \textit{et al.} \cite{chen2021intrinsic} learn disentangled shape and pose without supervision. However, the method of Zhou \textit{et al.} \cite{zhou2020unsupervised} can only train or test on registered data and the training process of Chen \textit{et al.} \cite{chen2021intrinsic} is very unstable and needs a great number of iterations to converge since the adoption of GAN \cite{goodfellow2020generative}. In this work, our method can work on non-registered meshes (they can have different vertex numbers and orders) in an unsupervised manner and enable a more robust and faster convergence. The cross consistency constraint mentioned in Zhou \textit{et al.} \cite{zhou2020unsupervised} is used on meshes with the same identity and is not sufficient to instruct the disentanglement. Different from it, our cross consistency learning scheme helps to achieve the unsupervised 3D pose transfer globally and effectively based on the main and auxiliary generators.

\section{Method}
\begin{figure*}
  \centering
  \includegraphics[scale=0.41]{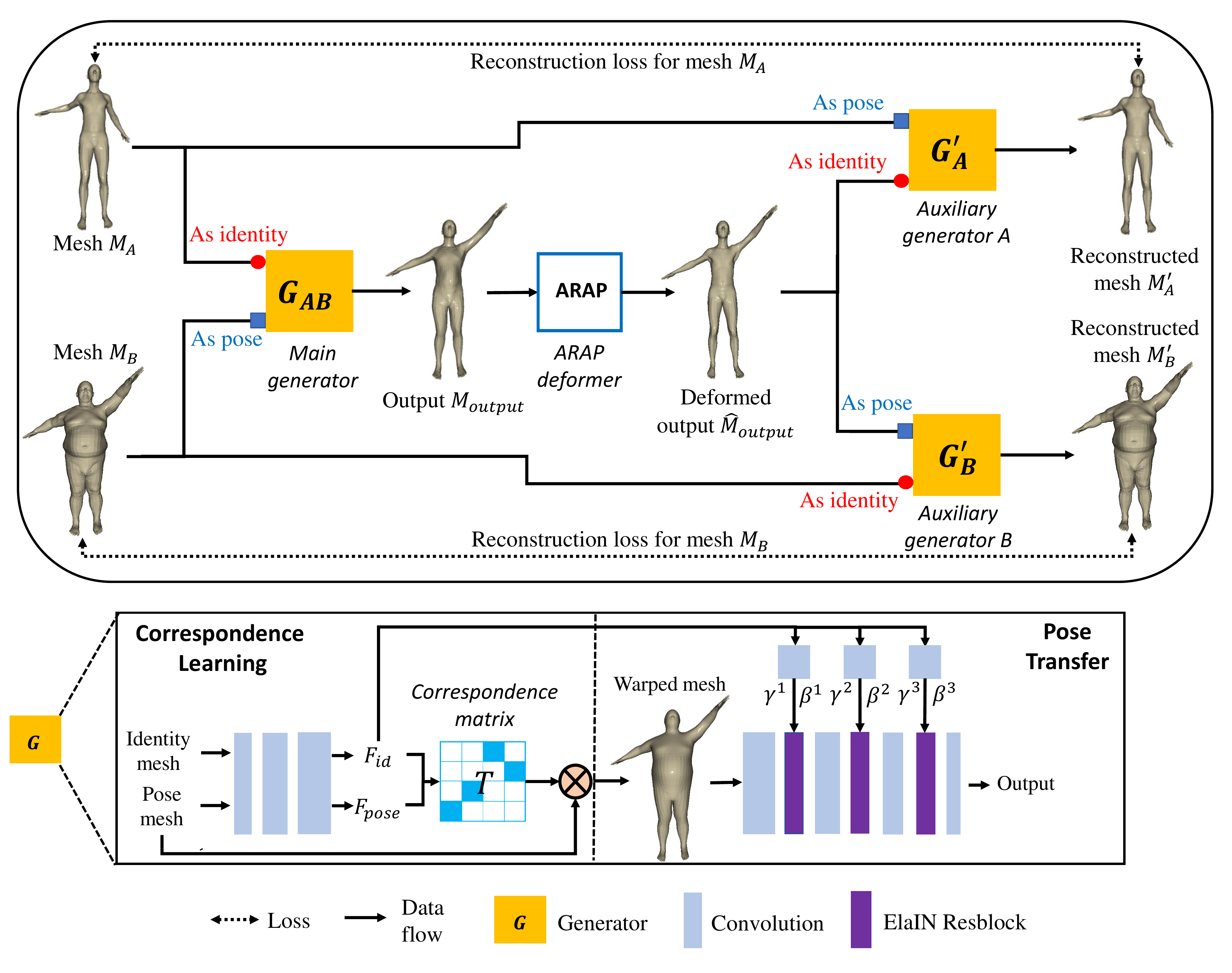}
  \caption{\textbf{An overview of \textit{X-DualNet}.} With a main generator $G_{AB}$, we take two meshes $M_{A}$ and $M_{B}$ as inputs and generate the output $M_{output} = G_{AB}(M_{A}, M_{B})$. Here, we view $M_{A}$ as the identity mesh and $M_{B}$ as the pose mesh respectively. We also add an ARAP deformer in the training process to fine-tune the body shape of the generated results and get the deformed output $\hat{M}_{output}$. Then we adopt an auxiliary generator $G^{\prime}_{A}$ to reconstruct the mesh $M_{A}$. Specifically, $G^{\prime}_{A}$ takes $M_{A}$ as the pose mesh and $\hat{M}_{output}$ as the identity mesh respectively to generate $M^{\prime}_{A} = G^{\prime}_{A}(\hat{M}_{output}, M_{A})$. Similarly, we can reconstruct $M_{B}$ with another auxiliary generator $G^{\prime}_{B}$ to generate $M^{\prime}_{B} = G^{\prime}_{B}(M_{B}, \hat{M}_{output})$. Note that $G_{AB}$, $G^{\prime}_{A}$ and $G^{\prime}_{B}$ share the same network parameters. The pose transfer procedure is consistent across the main generator and two auxiliary generators. We introduce a reconstruction loss to achieve the dual reconstruction objective $M^{\prime}_{A} \approx M_{A}$ and $M^{\prime}_{B} \approx M_{B}$. The generator $G$ contains two modules. In the correspondence learning module, we solve an optimal transport problem to build the correspondence between input meshes with the extracted features $F_{id}$ and $F_{pose}$. The pose transfer module contains several ElaIN residual blocks which help to achieve the 3D pose transfer.}
  \label{net}
\end{figure*}

With a source mesh and a target mesh, 3D pose transfer aims to transfer the pose from the source mesh to the target mesh while preserving the identity information (e.g., face, body shape) of the target mesh. In this section, we will introduce how to implement our \textit{X-DualNet} for unsupervised 3D pose transfer.

To explain our method better, we define a 3D triangle mesh as $M(I, P, N, O)$. Here, $I$ represents the identity of the mesh (e.g., face, body shape), $P$ means the pose of the mesh, $N$ and $O$ denote the vertex number and order respectively.


\subsection{Network architecture}
As shown in Figure \ref{net}, we adapt a generator $G$ as the basic component of our \textit{X-DualNet}. Given the identity mesh $M_{id} = M(I_{id}, P_{id}, N_{id}, O_{id})$ and the pose mesh $M_{pose} = M(I_{pose}, P_{pose}, N_{pose}, O_{pose})$, $G$ aims to transfer the pose from $M_{pose}$ to $M_{id}$ and generate the result $M^{\prime}(I_{id}, P_{pose}, N_{id}, O_{id})$. The inputs of $G$ are $V_{id} \in \mathbb{R}^{N_{id} \times 3}$ and $V_{pose} \in \mathbb{R}^{N_{pose} \times 3}$. Here, $N_{id}$ and $N_{pose}$ are the vertex number of $M_{id}$ and $M_{pose}$ respectively, $V_{id}$ and $V_{pose}$ are the corresponding vertex coordinates.

The basic component $G$ contains two modules: correspondence learning and pose transfer. In the correspondence learning module, we learn an optimal matching matrix between the identity mesh and the pose mesh using the optimal transport. We discuss more details in Section \ref{sec:correspondencelearning}. With the optimal matching matrix, the pose mesh is warped to get the warped mesh (as shown in Figure \ref{net}) which is actually a redistribution of the vertices of the pose mesh. 

The pose transfer module aims to generate the output which can combine the pose of $M_{pose}$ and the identity of $M_{id}$. This pose transfer procedure is achieved based on the proposed elastic instance normalization (ElaIN). We will introduce how to design ElaIN in Section \ref{sec:pose transfer}.



\subsubsection{Correspondence learning}
\label{sec:correspondencelearning}
Given an identity mesh and a pose mesh, our correspondence learning module calculates an optimal matching matrix, each element of the matching matrix represents the similarities between two vertices in the two meshes. The first step in our shape correspondence learning is to compute a correlation matrix with the extracted features, which is based on cosine similarity and denotes the matching similarities between any two positions from different meshes. However, the matching scores in the correlation matrix are calculated without any additional constraints. To learn a better matching, we solve this problem from a global perspective by modeling it as an optimal transport problem.

\noindent\textbf{Correlation matrix.}  
We first introduce our feature extractor which aims to extract features for the unordered input vertices. Different from NPT \cite{wang2020neural}, we use pointnet++-like convolution layers to extract vertex features.
Given the extracted vertex feature maps $f_{id} \in \mathbb{R}^{D \times N_{id}}$, $f_{pose} \in \mathbb{R}^{D \times N_{pose}}$ of identity and pose meshes ($D$ is the channel-wise
dimension), a popular method to compute correlation matrix is using the cosine similarity \cite{zhang2019deep, zhang2020cross}. Concretely, we compute the correlation matrix $\mathbf{C} \in \mathbb{R}^{N_{id} \times N_{pose}}$ as:
\begin{equation}
    \mathbf{C}(i, j) = \frac{f_{id}(i)^{\top}f_{pose}(j)}{\left\|f_{id}(i)\right\|\left\|f_{pose}(j)\right\|},
    \label{corr} 
\end{equation}
where $\mathbf{C}(i, j)$ denotes the individual matching score between $f_{id}(i)$ and $f_{pose}(j) \in \mathbb{R^{D}}$, $f_{id}(i)$ and $f_{pose}(j)$ represent the channel-wise feature of $f_{id}$ at position i and $f_{pose}$ at j. 

\noindent\textbf{Optimal transport problem.}
To learn a better matching with additional constraints in this work, we model our shape correspondence learning as an optimal transport problem. 
We first define a matching matrix $\mathbf{T} \in \mathbb{R}_{+}^{N_{id} \times N_{pose}}$ between identity and pose meshes. Then we can get the total correlation as $\sum_{i j}\mathbf{C}(i, j)\mathbf{T}(i, j)$. The aim will be maximizing the total correlation score to get an optimal matching matrix $\mathbf{T}_{m}$. 

We treat the correspondence learning between identity and pose meshes as the transport of mass. A mass that is equal to $N_{id}^{-1}$ will be assigned to each vertex in the identity mesh, and each vertex in the pose mesh will receive the mass $N_{pose}^{-1}$ from identity mesh through the built correspondence between vertices. Then if we define $\mathbf{Z} = 1 - \mathbf{C}$ as the cost matrix, our goal can be formulated as a standard optimal transport problem by minimizing the total matching cost,

\begin{equation}
\begin{aligned}
    \mathbf{T}_{m} &= \mathop{\arg\min}_{\mathbf{T} \in \mathbb{R}_{+}^{N_{id} \times N_{pose}}} \sum_{i j}\mathbf{Z}(i, j)\mathbf{T}(i, j)  \\
    s.t. \quad \mathbf{T}\mathbf{1}_{N_{pose}} &= \mathbf{1}_{N_{id}}N^{-1}_{id}, \quad \mathbf{T}^{\top}\mathbf{1}_{N_{id}} = \mathbf{1}_{N_{pose}}N^{-1}_{pose}.
\end{aligned}
    \label{tm}
\end{equation}

where $\mathbf{1}_{N_{id}} \in \mathbb{R}^{N_{id}}$ and $\mathbf{1}_{N_{pose}} \in \mathbb{R}^{N_{pose}}$ are vectors whose elements are all $1$. The first constraint in Equation \ref{tm} means that the mass of each vertex in $M_{id}$ will be entirely transported to some of the vertices in $M_{pose}$. And each vertex in $M_{pose}$ will receive a mass $N_{pose}^{-1}$ from some of the vertices in $M_{id}$ with the second constraint. This problem can be solved by the Sinkhorn-Knopp algorithm \cite{sinkhorn1967diagonal}. The details of the solving process will be given in the supplementary material.

With the matching matrix, we can warp the pose mesh and obtain the vertex coordinates $V_{warp} \in \mathbb{R}^{N_{id} \times 3}$ of the warped mesh,
\begin{equation}
    V_{warp}(i) = \sum_{j}\mathbf{T}_{m}(i,j)V_{pose}(j), 
    \label{warp}
\end{equation}
 the warped mesh $M_{warp}$ inherits the number and order of vertex from identity mesh and can be reconstructed with the face information of identity mesh as shown in Figure \ref{net}.
 
\subsubsection{Pose transfer}
\label{sec:pose transfer}
In this section, we introduce our pose transfer module which transfers the pose from the warped mesh to the identity mesh and generates the desired output. 

\noindent\textbf{Elastic instance normalization.}
Previous conditional normalization layers \cite{huang2017arbitrary, park2019semantic, wang2020neural} used in different tasks always calculated their denormalization parameters only with the external data. We argue that it may break the consistency and continuity of the original features. Inspired by Soft-AdaIN \cite{chen2020cartoonrenderer}, we propose \textit{Elastic Instance Normalization (ElaIN)} which blends the statistics of original features and the learned parameters from external data adaptively and elastically.

\begin{figure*}[h]
  \centering
  \includegraphics[scale=0.46]{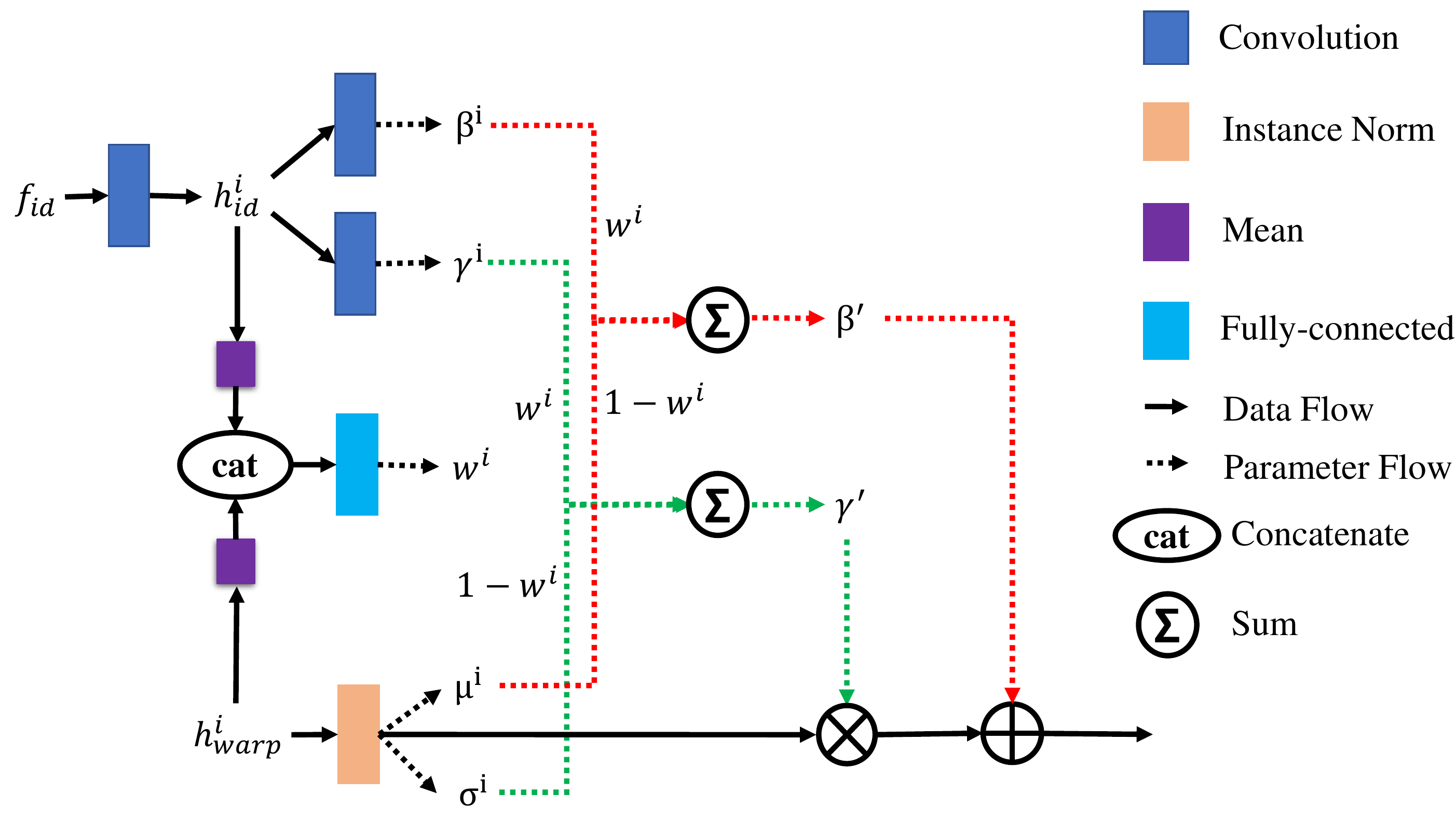}
  \caption{\textbf{The detailed design of our elastic instance normalization}. Here, we normalize the features of the warped mesh $h_{warp}^{i}$ with InstanceNorm and get the mean $\mu^{i}$ and standard deviation $\sigma^{i}$. Then, the features of the identity mesh are fed into a convolution layer to get $h_{id}^{i}$, which shares the same size with $h_{warp}^{i}$. We calculate the mean of $h_{warp}^{i}$, $h_{id}^{i}$ and concatenate them in channel dimension. A fully-connected layer is employed to compute an adaptive weight $w^{i}$. We blend $\gamma^{i}$, $\sigma^{i}$ and $\beta^{i}$, $\mu^{i}$ elastically with $w^{i}$ to get $\gamma^{\prime}$ and $\beta^{\prime}$. $\gamma^{i}$ and $\beta^{i}$ are learned from $h_{id}^{i}$. Finally, we scale the normalized $h_{warp}^{i}$ with $\gamma^{\prime}$ and shift it with $\beta^{\prime}$. The value on the parameter flow means the weight.}
  \label{elain}
\end{figure*}

To be specific, we let $h^{i} \in \mathbb{R}^{S^{i} \times D^{i} \times N^{i}}$ as the activation value before the $i\mbox{-th}$ normalization layer, where $S^{i}$ is the batch size, $D^{i}$ is the dimension of feature channel and $N^{i}$ is the number of vertex. 
As shown in Figure \ref{elain}, we normalize the feature maps of the warped mesh with instance normalization first and the mean and standard deviation are calculated across spatial dimension ($n \in N^{i}$) for each sample $s \in S^{i}$ and each channel $d \in D^{i}$,
\begin{equation}
    \mu^{i} = \frac{1}{N^{i}}\sum_{n}{h_{warp}^{i}},
\end{equation}
\begin{equation}
\label{sigma}
    \sigma^{i} = \sqrt{\frac{1}{N^{i}}\sum_{n}{(h_{warp}^{i} - \mu^{i})^{2}} + \epsilon},
\end{equation}
then the feature maps of the identity mesh are fed into a $1 \times 1$ convolution layer to get $h_{id}^{i}$, which shares the same size with $h_{warp}^{i}$. We calculate the mean of $h_{warp}^{i}$, $h_{id}^{i}$ to make them $S^{i} \times D^{i} \times 1$ tensors. The tensors are then concatenated in channel dimension to get a $S^{i} \times (2D^{i}) \times 1$ tensor. A fully-connected layer is employed to compute an adaptive weight $w(h_{warp}^{i}, h_{id}^{i}) \in \mathbb{R}^{S^{i} \times D^{i} \times 1}$ with the concatenated tensor. With $w(h_{warp}^{i}, h_{id}^{i})$, we can define the modulation parameters of our normalization layer.
\begin{equation}
    \begin{split}
        \gamma^{\prime}(h_{warp}^{i}, h_{id}^{i}) = w(h_{warp}^{i}, h_{id}^{i})\gamma^{i} + (1 - w(h_{warp}^{i}, h_{id}^{i}))\sigma^{i}, \\
    \beta^{\prime}(h_{warp}^{i}, h_{id}^{i}) = w(h_{warp}^{i}, h_{id}^{i})\beta^{i} + (1 - w(h_{warp}^{i}, h_{id}^{i}))\mu^{i}.    
    \end{split}
\end{equation}
where $\gamma^{i}$ and $\beta^{i}$ are learned from the identity feature $h_{id}^{i}$ with two convolution layers. Finally, we can scale the normalized $h_{warp}^{i}$ with $\gamma^{\prime}$ and shift it with $\beta^{\prime}$,

\begin{equation}
    \begin{aligned}
       \mbox{ElaIN}(h_{warp}^{i}, h_{id}^{i}) = &\gamma^{\prime}(h_{warp}^{i}, h_{id}^{i})(\frac{h_{warp}^{i} - \mu^{i}}{\sigma^{i}}) \\
       &+ \beta^{\prime}(h_{warp}^{i}, h_{id}^{i}). 
       \end{aligned}
       \label{soft}
\end{equation}
\noindent\textbf{Pose transfer module.}
Our pose transfer module is designed to transfer the pose from the warped mesh to the identity mesh. Following NPT \cite{wang2020neural}, CoCosNet \cite{zhang2020cross}, and SPADE\cite{park2019semantic}, we design the ElaIN residual block with our ElaIN in the form of ResNet blocks \cite{he2016identity}. As shown in Figure \ref{net}, our architecture contains 3 ElaIN residual blocks. Each of them consists of our proposed ElaIN followed by a simple convolution layer and LeakyReLU. With the ElaIN residual blocks, our network can generate high-quality results.

\subsection{Cross consistency and dual reconstruction}
\label{indu}

In this section, we will introduce how to achieve unsupervised 3D pose transfer with our cross consistency learning scheme and dual reconstruction objective.

As shown in Figure \ref{net}, with a main generator $G_{AB}$, we take two meshes $M_{A}$ and $M_{B}$ as inputs and generate the output $M_{output} = G_{AB}(M_{A}, M_{B})$. Here, we view $M_{A}$ as the identity mesh and $M_{B}$ as the pose mesh respectively. We also use an ARAP deformer that will be introduced in the following to get the deformed output $\hat{M}_{output}$. Then we adopt an auxiliary generator $G^{\prime}_{A}$ to reconstruct the mesh $M_{A}$. Specifically, $G^{\prime}_{A}$ takes $M_{A}$ as the pose mesh and $\hat{M}_{output}$ as the identity mesh respectively to generate $M^{\prime}_{A} = G^{\prime}_{A}(\hat{M}_{output}, M_{A})$. Similarly, we can reconstruct $M_{B}$ with another auxiliary generator $G^{\prime}_{B}$. For $G^{\prime}_{B}$, we view $M_{B}$ as the identity mesh and $\hat{M}_{output}$ as the pose mesh respectively to generate $M^{\prime}_{B} = G^{\prime}_{B}(M_{B}, \hat{M}_{output})$. Note that $G_{AB}$, $G^{\prime}_{A}$ and $G^{\prime}_{B}$ share the same network parameters. The pose transfer procedure is consistent across the main generator and two auxiliary generators.

To achieve the dual reconstruction objective, we introduce a reconstruction loss in the loss function section that encourages $M^{\prime}_{A} \approx M_{A}$ and $M^{\prime}_{B} \approx M_{B}$. Once the dual reconstruction objective is achieved, our pose transfer module will not only learn the identity information from the identity mesh but also inherit the pose from the pose mesh. 

Based on this cross consistency learning scheme and the dual reconstruction objective, our \textit{X-DualNet} can learn the 3D pose transfer without any supervision of the ground truth label.

\subsection{ARAP deformer}
To keep the original body shape of the generated results $M_{output}$, we adopt an as-rigid-as-possible (ARAP) deformer in the training process as shown in Figure \ref{net},
\begin{equation}
    \hat{M}_{output} = \mbox{ARAP} (M_{A}, M_{output}).
\end{equation}
Here, $M_{A}$ is the identity mesh for the generation of $M_{output}$. With the ARAP deformer, we deform $M_{A}$ to match a few randomly fixed vertices of the generated mesh $M_{output}$. As shown in Figure \ref{net}, the deformed output $\hat{M}_{output}$ has better preservation of the body shape of the original identity mesh.

In the following, we introduce how ARAP works to preserve the body shape of the meshes. We define $M$ as the original mesh and $\hat{M}$ as the deformed mesh. For the cell $C_{i}$ (It includes vertex $v_{i}$ and its one-ring neighbor $\mathcal{N}(i)$.) corresponding to vertex $v_{i}$ and the deformed cell $\hat{C}_{i}$, if the deformation from $C_{i}$ to $\hat{C}_{i}$ is rigid, then there is a rotation matrix $\mathbf{R}_{i}$ satisfies,
\begin{equation}
    \hat{v}_{i} - \hat{v}_{j} = \mathbf{R}_{i}(v_{i} - v_{j}),   \quad \forall j \in \mathcal{N}(i).
\end{equation}

If the deformation between $C_{i}$ and $\hat{C}_{i}$ is not rigid, we can still find the optimal rotation matrix that makes the deformation from $C_{i}$ to $\hat{C}_{i}$ as rigid as possible, i.e., minimizes the energy
\begin{equation}
   E(C_{i}, \hat{C}_{i}) = \sum_{j \in \mathcal{N}(i)}w_{ij}\left\|(\hat{v}_{i} - \hat{v}_{j}) - \mathbf{R}_{i}(v_{i} - v_{j})\right\|^{2},
   \label{arap}
\end{equation}
where $w_{ij}$ means the per-edge weight. In this work, we use the cotangent weighting for $w_{ij}$. ARAP uses an alternating minimization strategy to solve Equation \ref{arap}. For a given set of vertex positions $\hat{v}$, it calculates the optimal rigid transformations $\{\mathbf{R}_{i}\}$ that minimize $E$. Then it updates the vertex positions $\hat{v}$ with the calculated $\{\mathbf{R}_{i}\}$ while minimizing $E$. These iterations continue until the local energy minimum is reached. For the detailed solving procedure please refer to \cite{sorkine2007rigid}.

In this work, we only add the ARAP deformer in the training loop to help keep the body shape of $M_{output}$. Different from Zhou \textit{et al.} \cite{zhou2020unsupervised}, we adopt ARAP on non-registered data. Under this circumstance, we have a trade-off between the training efficiency and the number of epochs that adds ARAP deformer. Through extensive experiments, we finally decided to add the ARAP deformer in the last 50 epochs for training. 

\subsection{Loss functions}

\subsubsection{Reconstruction loss.}

Since our network is trained without the supervision of the ground truth label, we introduce a reconstruction loss to achieve the dual reconstruction objective in Section \ref{indu}. The reconstruction loss is calculated using the $L2$ distance between the vertex coordinates of the reconstructed meshes and the original meshes. For the reconstruction of $M_{A}$,
\begin{equation}
    \mathcal{L}_{A\_rec} = \left\|V^{\prime}_{A} - V_{A}\right\|_{2}^{2},
\end{equation}
where $V^{\prime}_{A}$ and $V_{A} \in \mathbb{R}^{N_{A} \times 3}$ are the vertex coordinates of $M^{\prime}_{A} = G^{\prime}_{A}(\hat{M}_{output}, M_{A})$ and $M_{A}$ respectively. $N_{A}$ is the vertex number. Similarly,
\begin{equation}
    \mathcal{L}_{B\_rec} = \left\|V^{\prime}_{B} - V_{B}\right\|_{2}^{2}, 
\end{equation}
where $V^{\prime}_{B}$ and $V_{B} \in \mathbb{R}^{N_{B} \times 3}$ are the vertex coordinates of $M^{\prime}_{B} = G^{\prime}_{B}(M_{B}, \hat{M}_{output})$ and $M_{B}$ respectively. $N_{B}$ is the vertex number.

With the reconstruction loss, we can achieve the 3D pose transfer without any supervision and the meshes generated by our model are very close to the ground truth.

\subsubsection{Backward correspondence loss.}

To learn better shape correspondence in the correspondence learning module, we also introduce a backward correspondence loss inspired by CorrNet3D \cite{zeng2021corrnet3d}. As discussed in Section \ref{sec:correspondencelearning}, we define the optimal matching matrix as $\mathbf{T}_{m} \in \mathbb{R}_{+}^{N_{id} \times N_{pose}}$ between the identity and pose meshes. With $\mathbf{T}_{m}$, we can warp the pose mesh and obtain the warped pose mesh $M_{warp}$.

Then we can reconstruct the pose mesh based on the transpose of the correspondence matrix and the warped mesh,
\begin{equation}
    V^{\prime}_{pose}(i) = \sum_{j}\mathbf{T}_{m}^{\top}(i,j)V_{warp}(j) \label{warp}.
\end{equation}
Here, $i \in [0, N_{pose})$ and $j \in [0, N_{id})$. Finally, the backward correspondence loss can be defined as,
\begin{equation}
\begin{aligned}
    \mathcal{L}_{corr} &= \left\|V^{\prime}_{pose} - V_{pose}\right\|_{2}^{2} \\
     & = \left\|\mathbf{T}_{m}^{\top}\mathbf{T}_{m}V_{pose} - V_{pose}\right\|_{2}^{2},
\end{aligned}
\label{lcorr}
\end{equation}
we can also rewrite the backward correspondence loss as,
\begin{equation}
    \mathcal{L}_{corr} = \left\|\mathbf{T}_{m}^{\top}\mathbf{T}_{m} - \mathbf{I}_{N_{pose}}\right\|_{2}^{2},
    \label{lcorr2}
\end{equation}
where $\mathbf{I}_{N_{pose}}$ is the identity matrix whose size is $N_{pose} \times N_{pose}$. Minimizing Equation \ref{lcorr} is equivalent to minimizing Equation \ref{lcorr2}. 

\subsubsection{Edge loss.}

Following several 3D mesh generation works \cite{pan2019deep, wang2020neural, wang2018pixel2mesh}, we adopt edge loss which can penalize flying vertices and make the connection between vertices smoother. Since the reconstruction loss does not consider the vertex connection, the predicted mesh can be affected by flying vertices, which usually leads to excessively long edges. Edge loss can help alleviate this issue. We can define the edge loss as,
\begin{equation}
    \mathcal{L}_{edge} = \sum_{v}{\sum_{v_{\mathcal{N}} \in \mathcal{N}(v)}{\left\|v - v_{\mathcal{N}}\right\|_{2}^{2}}},
\end{equation}
where $v \in V_{output}$ is the vertex of the output mesh and $\mathcal{N}(v)$ is the neighbor of $v$.

To sum up, the total loss $\mathcal{L}$ for unsupervised 3D pose transfer is,
\begin{equation}
    \mathcal{L} = \lambda_{rec}(\mathcal{L}_{A\_rec} + \mathcal{L}_{B\_rec})+ \lambda_{corr}\mathcal{L}_{corr} + \mathcal{L}_{edge},
\end{equation}
where $\lambda_{rec}$, $\lambda_{corr}$ denote the weight of reconstruction loss and backward correspondence loss respectively.

\section{Experiments}
\label{experiment}
\noindent\textbf{Datasets.}
We use the same human mesh dataset generated by SMPL \cite{loper2015smpl} as NPT \cite{wang2020neural} and 3D-CoreNet \cite{song20213d}. 
This dataset consists of 30 identities with 800 poses. Each mesh in the dataset has 6890 vertices. For the training set, we randomly choose 6400 mesh pairs (identity and pose meshes) and shuffle them every epoch. All vertices of the meshes will be shuffled randomly before input to be close to the real-world problem. And we also shift the meshes to the center according to their bounding boxes. When doing the test, we randomly choose 400 unseen mesh pairs that are different from 3D-CoreNet \cite{song20213d} for evaluation. They are pre-processed in the same manner as the training data. To further verify the generalization capability of our approach, we also test it on FAUST \cite{bogo2014faust} and MGN \cite{bhatnagar2019multi} in the experiments.

We use the animal mesh dataset generated by SMAL model \cite{zuffi20173d} in this work. It includes 21 Felidae animals, 5 Canidae animals, 8 Equidae animals, 4 Bovidae animals, and 3 Hippopotamidae animals. Each mesh has 3889 vertices. For the training set, we randomly choose 9000 mesh pairs from 21 identities (10 Felidae animals, 3 Canidae animals, 4 Equidae animals, 2 Bovidae animals, and 2 Hippopotamidae animals) with 400 poses. And we randomly choose 400 unseen pairs for testing. The choice of testing data is also different from 3D-CoreNet \cite{song20213d}. The animal meshes are pre-processed in the same way as we do in the human dataset.

\noindent\textbf{Evaluation metrics.} 
To compare different methods quantitatively, we use Pointwise Mesh Euclidean Distance (PMD), Chamfer Distance (CD), and Earth Mover's Distance (EMD) as our evaluation metrics. For all of them, the lower is better. We compare the averages of these three metrics obtained on the test datasets for quantitative evaluation.

\noindent\textbf{Implementation details.}
The values of hyperparameters in the loss function are $\lambda_{rec} = 2000$, $\lambda_{corr} = 200$. Our model is trained for 200 epochs with the Adam optimizer \cite{kingma2014adam} on two RTX 3090 GPUs, the learning rate is initialized as $1e\mbox{-}4$ in the first 100 epochs and decays $1e\mbox{-}6$ each epoch from the 100th epoch. For more implementation details of the network, please refer to our supplemental material.

\subsection{Baselines}
\noindent\textbf{Neural pose transfer.} Neural pose transfer(NPT) \cite{wang2020neural} transfers the pose with spatially adaptive instance normalization (SPAdaIN). It does not consider any correspondence between different meshes. Therefore, the performance of the model will be degraded although their method is convenient. We train NPT using the implementations provided by the authors.

\noindent\textbf{3D-CoreNet.} We can achieve a supervised 3D pose transfer method, which is named 3D-CoreNet following \cite{song20213d}. 3D-CoreNet contains a single generator $G$ and is trained with the supervision of the ground truth mesh $M_{gt}$. We first process the ground truth mesh to have the same vertex order as the identity mesh. Then we define a new reconstruction loss by calculating the point-wise $L2$ distance between the vertices of $M_{output}$ and $M_{gt}$, 
\begin{equation}
    \mathcal{L}_{rec} = \left\|V_{output} - V_{gt}\right\|_{2}^{2}
\end{equation}
where $V_{output}$ and $V_{gt} \in \mathbb{R}^{N_{id} \times 3}$ are the vertices of $M_{output}$ and $M_{gt}$ respectively. Notice that they all share the same size and order with the vertices of the identity mesh. With the reconstruction loss, the mesh predicted by 3D-CoreNet will be closer to the ground truth. The inference of 3D-CoreNet and X-DualNet only needs a single generator.

\begin{table*}
  \centering
  \caption{\textbf{Quantitative comparison.} We compare X-DualNet with NPT, 3D-CoreNet, and the proposed unsupervised baseline on human and animal meshes. For the evaluation metrics, the lower is better. 3D-CoreNet has the best performance. X-DualNet achieves comparable performances as 3D-CoreNet and even outperforms NPT which is a fully supervised method.}
  \label{comparison}
  \begin{tabular}{cccccc}
    \toprule
  Dataset  & Type & Method &  PMD ($\times10^{-3}$)  & CD ($\times10^{-3}$) & EMD ($\times10^{-2}$) \\
    \midrule
  \multirow{4}{*}{SMPL} & \multirow{2}{*}{\textit{Sup.}} &  NPT  & 2.77 & 5.91 & 7.88\\
    &  &  3D-CoreNet  & \textbf{0.49}  &  \textbf{0.90}  &   \textbf{3.61}  \\ \cmidrule{2-6} 
     & \multirow{2}{*}{\textit{Unsup.}}&  Baseline   &  2.23 & 4.05  &  7.05   \\
     &   &  Ours   & 1.67 & 2.71 & 5.96      \\
    \midrule
    \multirow{4}{*}{SMAL} & \multirow{2}{*}{\textit{Sup.}} &NPT  & 8.01 & 17.76 & 12.57 \\
     &  & 3D-CoreNet  & \textbf{3.55}  &  \textbf{6.49}  &\textbf{9.09}     \\ \cmidrule{2-6} 
         & \multirow{2}{*}{\textit{Unsup.}} &  Baseline   & 16.50  & 37.54  &  23.62   \\
      &  & Ours  & 4.48 & 9.05 & 10.19     \\
    \bottomrule
  \end{tabular}
\end{table*}

\begin{figure*}
  \centering
  \includegraphics[scale=0.37]{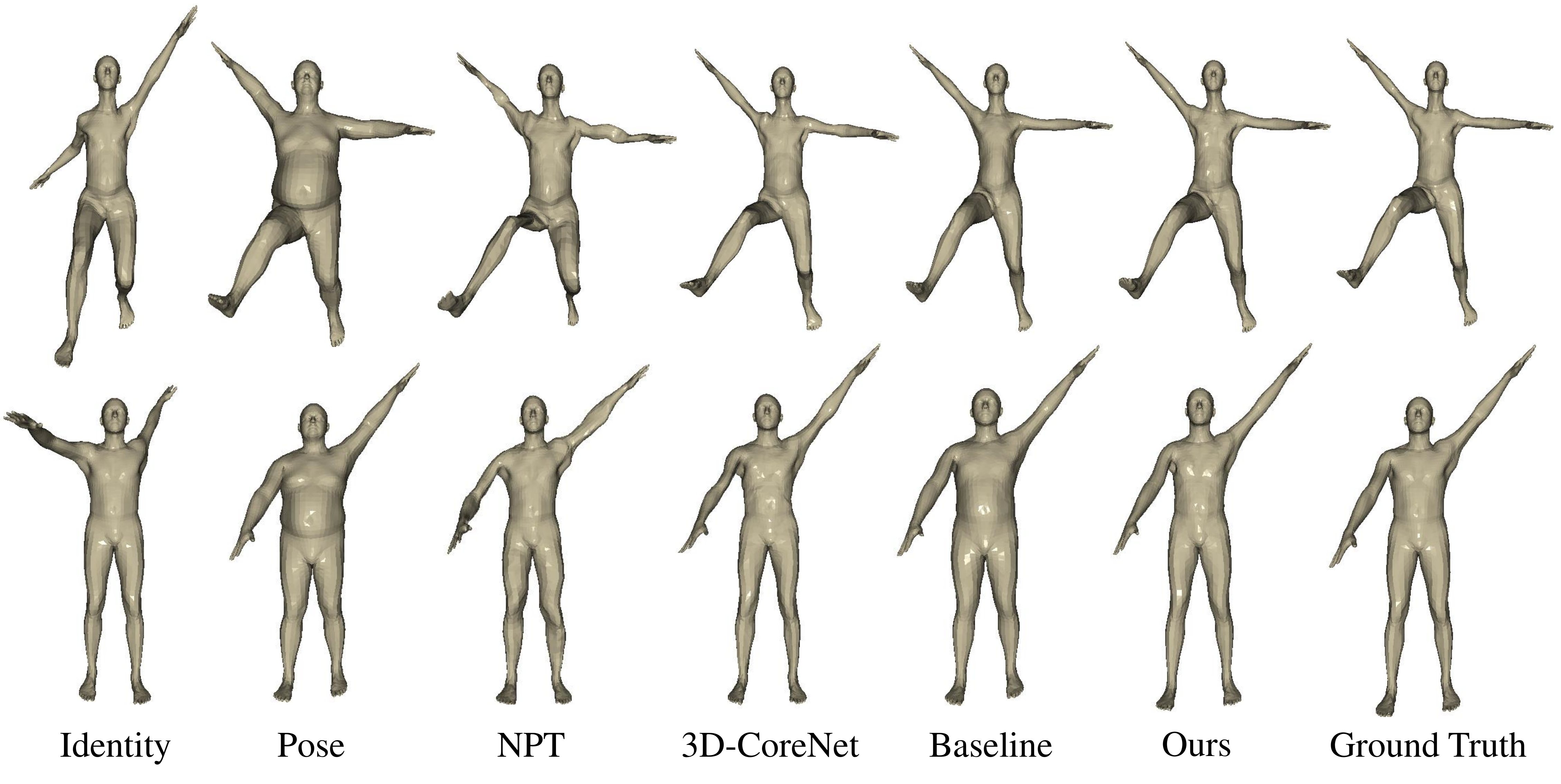}
  \caption{\textbf{Qualitative comparison on human data.} Our method and 3D-CoreNet can generate better results than NPT and the proposed unsupervised baseline. The surfaces of meshes generated by NPT are not always smooth. And the proposed unsupervised baseline cannot preserve the body shapes very well. The human meshes are from SMPL \cite{loper2015smpl}.}
  \label{SMPL}
\end{figure*}

\begin{figure*}
  \centering
  \includegraphics[scale=0.22]{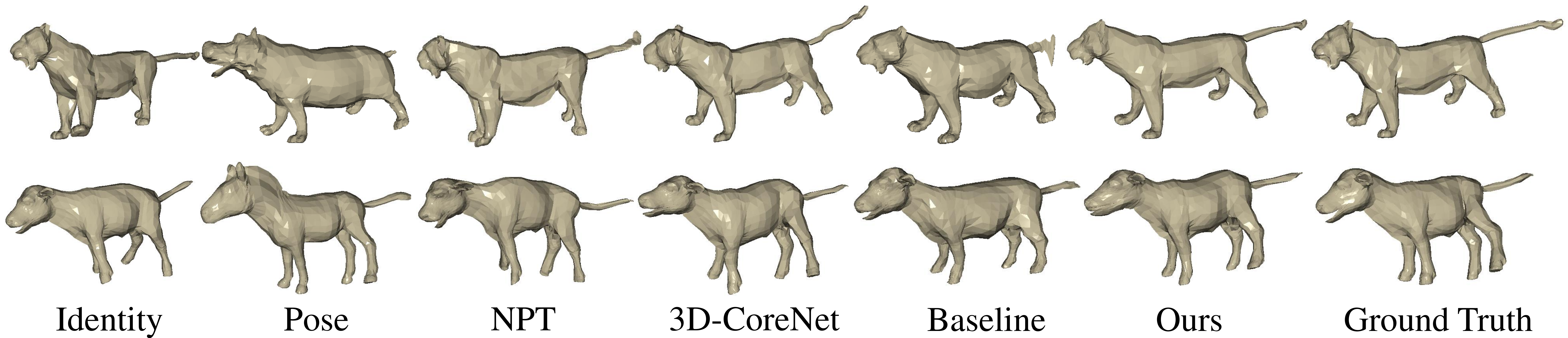}
  \caption{\textbf{Qualitative comparison on animal data.} Our method and 3D-CoreNet produce satisfactory results on the animal data. NPT produces many artifacts and cannot transfer the pose successfully. The proposed baseline sometimes cannot preserve the shape identity (e.g., the tail) well. The animal meshes are from SMAL \cite{zuffi20173d}.}
  \label{smal}
\end{figure*}

\noindent\textbf{Unsupervised baseline.}
We design the unsupervised baseline into two modules: cycle reconstruction and self reconstruction. It contains five generators. All five generators share the same network parameters. And we adopt three reconstruction losses to instruct the training. Compared with \textit{X-DualNet}, the unsupervised baseline needs more time for training. Please refer to the appendices for the detailed structure of the proposed unsupervised baseline. 

\begin{figure}[t]
  \centering
  \includegraphics[scale=0.28]{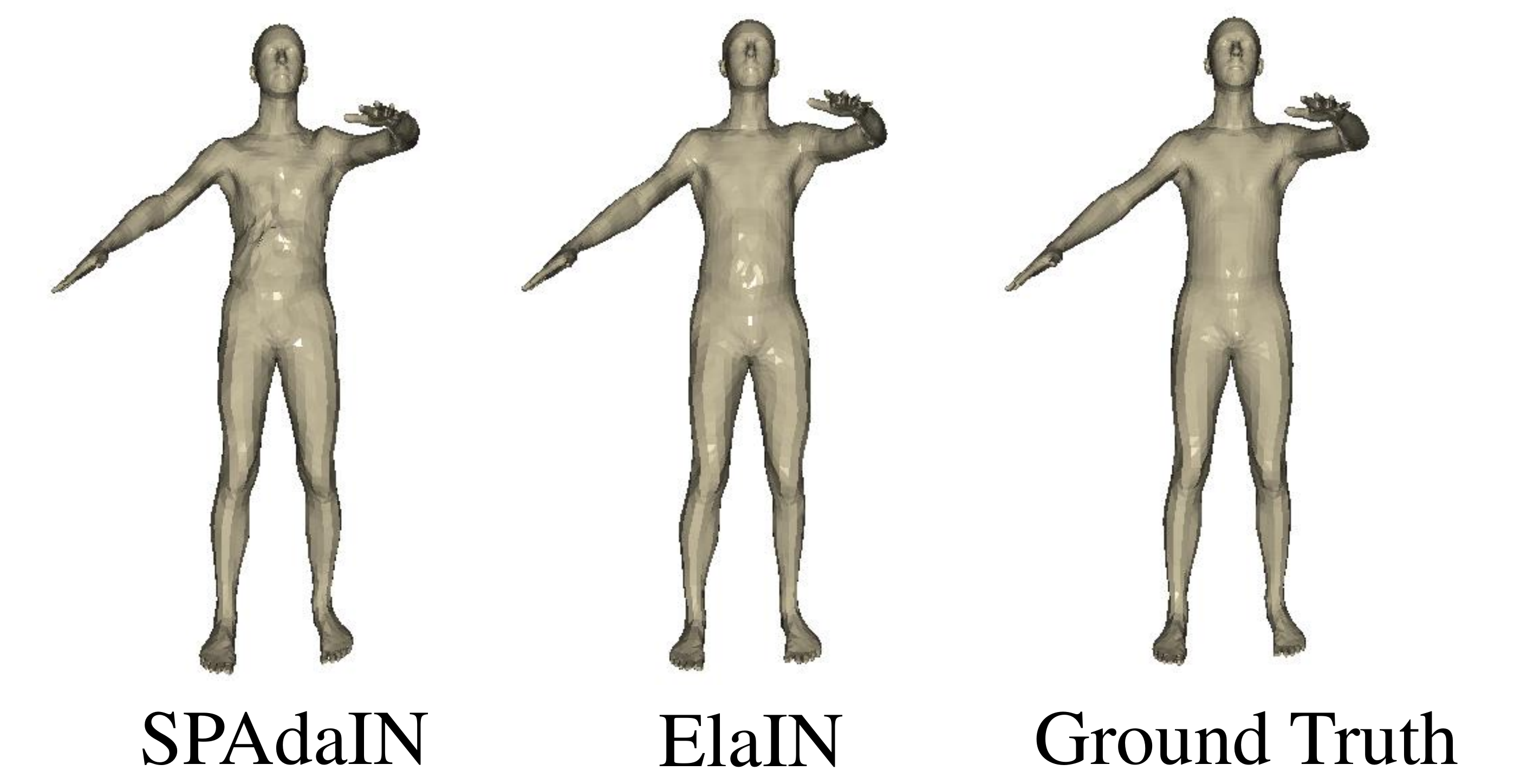}
  \caption{\textbf{Comparison of different conditional normalization layers.} We compare our ElaIN with SPAdaIN used in NPT \cite{wang2020neural} on SMPL \cite{loper2015smpl}. The surface of the mesh has clear artifacts and is not smooth when we replace ElaIN with SPAdaIN. Experiments are carried out on 3D-CoreNet.}
  \label{elain_spadain}
\end{figure}

\subsection{Comparison with the state-of-the-art approaches}

\begin{figure*}[b]
  \centering
  \includegraphics[scale=0.3]{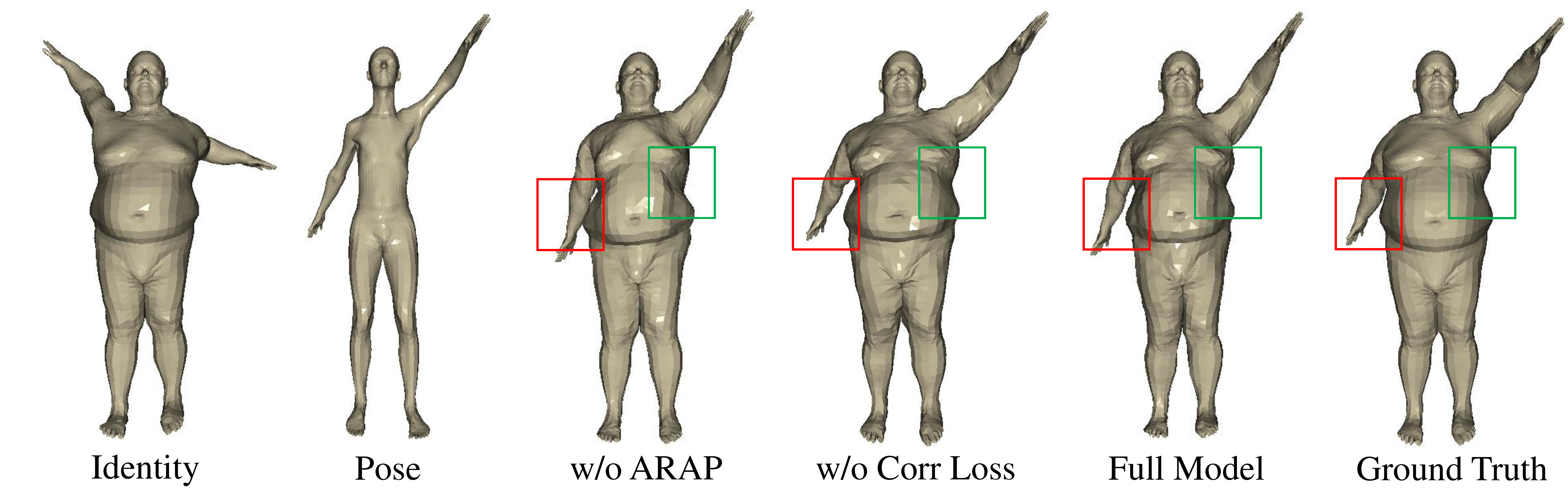}
  \caption{\textbf{Qualitative ablation studies of two components in X-DualNet.} When we do not add the ARAP deformer in the training loop, the generated results do not preserve the body shape well which can be shown in the green bounding boxes. In the third column, the body shape is sunken from the left and right sides. When we remove the backward correspondence loss $\mathcal{L}_{corr}$ in the training, the pose transfer results are not accurate which can be shown in the red bounding boxes. In the fourth column, the right arm is farther from the body than others. And the reason why the right arm of the mesh in the third column looks so close to the body is that the body shape is not well maintained without ARAP deformer. The input human meshes are from SMPL \cite{loper2015smpl}. w/o means we remove this component.}
  \label{ablation}
\end{figure*}

We compare \textit{X-DualNet} with NPT \cite{wang2020neural}, 3D-CoreNet, and our proposed unsupervised baseline. NPT and 3D-CoreNet are supervised with the ground truth. 

Figure \ref{SMPL} provides the qualitative results tested on SMPL \cite{loper2015smpl}. It can be clearly seen that our method and 3D-CoreNet can produce better results with the least artifacts. The surfaces of meshes generated by NPT are not smooth, especially on the legs and arms because NPT does not learn the shape correspondence. The proposed unsupervised baseline cannot preserve the body shapes well although it transfers the poses successfully. To be clear, we will discuss the reason in supplemental material following the detailed structure of the unsupervised baseline. The qualitative results tested on animal data are shown in Figure \ref{smal}, NPT always fails to transfer the pose and produces many artifacts. The proposed baseline sometimes cannot keep the identity information (e.g., the tail). In comparison, our method and 3D-CoreNet still produce successful results. 

To compare different methods quantitatively, we use PMD, CD and EMD as the evaluation metrics which are calculated from the ground truth and the generated meshes. We show the comparison results in Table \ref{comparison}. 3D-CoreNet has the best performance. Our \textit{X-DualNet} outperforms NPT and the proposed unsupervised baseline in all metrics over two datasets although NPT is trained with the ground truth.

\begin{figure*}
  \centering
  \includegraphics[scale=0.3]{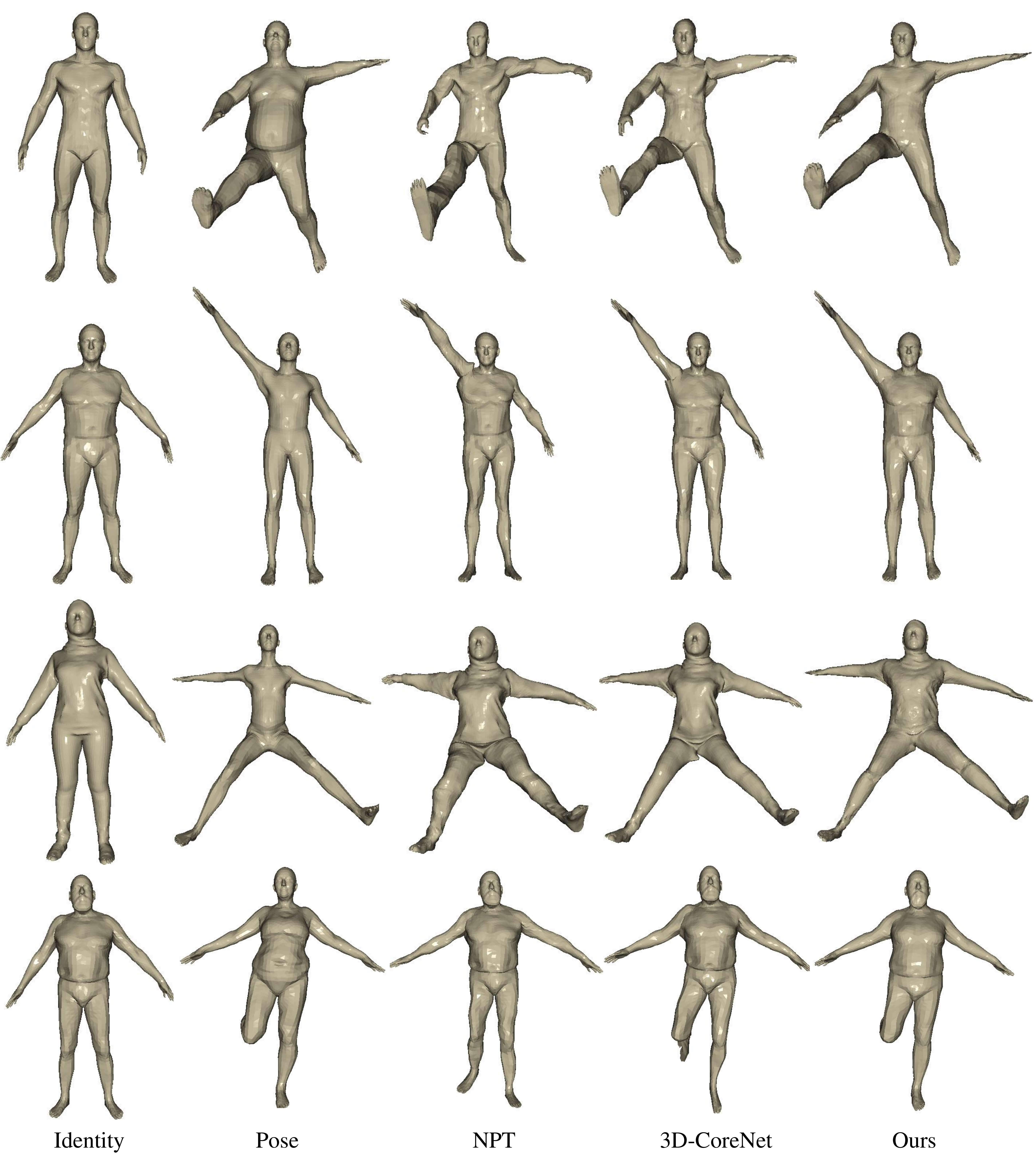}
  \caption{\textbf{Pose transfer results on FAUST \cite{bogo2014faust} and MGN \cite{bhatnagar2019multi}.} We choose four pose meshes and transfer the poses to the unseen meshes in FAUST and MGN. To evaluate the generalization ability of X-DualNet, we compare it with NPT \cite{wang2020neural} and 3D-CoreNet. As shown in the first three rows, the results generated by NPT and 3D-CoreNet always have many artifacts on the mesh surfaces. Our results are smoother than theirs. In the last row, NPT does not transfer the pose successfully since they do not learn the shape correspondence. 3D-CoreNet also fails to transfer the pose accurately. Our method has the best performance when meeting unseen poses. More generated results will be given in the supplemental material.}
  \label{generalization}
\end{figure*}

\subsection{Ablation studies}
In this section, we study the effectiveness of several components in our model on human data.

\noindent\textbf{ElaIN.} We compare our ElaIN with SPAdaIN proposed in NPT \cite{wang2020neural} for 3D pose transfer to verify the effectiveness of ElaIN.
To make the comparisons clearer, we conduct experiments on 3D-CoreNet, which requires only one generator since ElaIN works in the pose transfer module. When we replace our ElaIN with SPAdaIN, the surface of the mesh has clear artifacts and is not smooth as shown in Figure \ref{elain_spadain}. The metrics are also worse than using ElaIN as shown in Table \ref{elain_spadain_1}. We can know that ElaIN is very helpful in generating high-quality results.

\begin{table}[t]
  \caption{\textbf{Comparison of different conditional normalization layers.} We compare our ElaIN with SPAdaIN used in NPT \cite{wang2020neural}. The lower is better. Experiments are carried out on 3D-CoreNet.}
  \label{elain_spadain_1}
  \centering
  \begin{tabular}{ccccccc}
  \hline
    \toprule
       Dataset &  & SPAdaIN & ElaIN \\
    \midrule
       &PMD     &  1.01  &  \textbf{0.49}  \\
  SMPL & CD   & 2.13  & \textbf{0.90}      \\
   & EMD     & 4.66   &  \textbf{3.61} \\
    \bottomrule
  \end{tabular}
\end{table}

\noindent\textbf{ARAP deformer.}
We evaluate our \textit{X-DualNet} without the as-rigid-as-possible (ARAP) deformer. The qualitative ablation study results are shown in Figure \ref{ablation}. When we do not add the 
\begin{table}[h]
\centering
\caption{\textbf{Quantitative ablation studies of two components in X-DualNet.} We evaluate the importance of ARAP deformer and the backward correspondence loss. For PMD, CD, and EMD, the lower is better. w/o means we remove the component.}
      \label{ablation_1}
  \begin{tabular}{ccccc}
  \hline
    \toprule
       Dataset &   &  PMD & CD  & EMD \\
    \midrule
       & w/o ARAP    & 1.98 & 3.37 & 6.60     \\
  SMPL &w/o $\mathcal{L}_{corr}$       &  1.84 & 3.20  & 6.35  \\
   &Full model      & \textbf{1.67}  & \textbf{2.71} &  \textbf{5.96} \\
    \bottomrule
  \end{tabular}
\end{table}
ARAP deformer in the training loop, the generated results do not preserve the body shape well which can be shown in the green bounding boxes. In the third column, the body shape is sunken from the left and right sides. The quantitative ablation study analysis is shown in Table \ref{ablation_1}. As we can see, the quantitative performance of our method degrades when removing the ARAP deformer.

\noindent\textbf{Backward correspondence loss.}
We test the effectiveness of the backward correspondence loss $\mathcal{L}_{corr}$ in \textit{X-DualNet}. When we remove $\mathcal{L}_{corr}$ in the training, the pose transfer results will not be accurate. As we can see from the red bounding boxes in Figure \ref{ablation}, the poses of the generated results that add the backward correspondence loss are more similar to the ground truth. In the fourth column, the right arm is farther from the body than others. And the reason why the right arm of the mesh in the third column looks so close to the body is that the body shape is not well maintained without ARAP deformer. The quantitative results are also worse after removing the backward correspondence loss. We will visualize the built shape correspondence between human or animal meshes in the supplementary materials. 

\subsection{Generalization capability}

We evaluate \textit{X-DualNet} on unseen datasets FAUST \cite{bogo2014faust} and MGN \cite{bhatnagar2019multi} to verify the generalization capability of it. In FAUST, the human meshes have more unseen identities which share the same number of vertices with SMPL \cite{loper2015smpl}. For human meshes in MGN, they are dressed in clothes and they all have 27,554 vertices.

As shown in Figure \ref{generalization}, we choose four pose meshes and transfer the poses to the human meshes in FAUST and MGN. Our method still produces high-quality results when testing on unseen datasets although the number of vertices is different. To further verify the generalization ability of \textit{X-DualNet}, we compare it with NPT \cite{wang2020neural} and 3D-CoreNet. As shown in the first three rows, the results generated by NPT and 3D-CoreNet always have many artifacts on the mesh surfaces. Our results are smoother than both of them. In the last row, the input identity mesh and the pose mesh are both from FAUST. NPT does not transfer the pose successfully since they do not consider the correspondence between the identity and pose meshes. 3D-CoreNet also fails to transfer the pose accurately. Our unsupervised method has better performance when meeting unseen poses. 

\section{Conclusion}

In this work, we introduce \textit{X-DualNet}, a simple yet effective approach that enables unsupervised 3D pose transfer. In \textit{X-DualNet}, we propose a generator $G$ which contains correspondence learning and pose transfer modules to achieve the 3D pose transfer. We learn the shape correspondence by solving an optimal transport problem without any key point annotations and generate high-quality final meshes with our proposed elastic instance normalization (ElaIN) in the pose transfer module. To the best of our knowledge, our method is the first to learn the correspondence between different meshes and transfer the poses jointly in the 3D pose transfer task.

With the generator $G$ as the basic component, we propose a cross consistency learning scheme and a dual reconstruction objective to learn the pose transfer without any supervision of the ground truth label. Besides that, we also adopt an as-rigid-as-possible (ARAP) deformer to help preserve the body shape of the generated results. Extensive experiments on human and animal meshes demonstrate that our unsupervised solution achieves comparable performance as the state-of-the-art supervised approaches qualitatively and quantitatively and even outperforms some of them. In addition, our method has better generalization capability than supervised methods. In the future, we will try to build a system to connect the pose transfer problems between 2D images and 3D representations.

\ifCLASSOPTIONcompsoc
  \section*{Acknowledgments}
\else
  \section*{Acknowledgment}
\fi

This research is supported under the RIE2020 Industry Alignment Fund – Industry Collaboration Projects (IAF-ICP) Funding Initiative, as well as cash and in-kind contribution from the industry partner(s). This research is also supported by the National Research Foundation, Singapore under its AI Singapore Programme (AISG Award No: AISG-RP-2018-003), the Ministry of Education, Singapore, under its Academic Research Fund Tier 2 (MOE-T2EP20220-0007) and Tier 1 (RG95/20). This research is also supported by the Agency for Science, Technology and Research (A*STAR), Singapore under its MTC Young Individual Research Grant (Grant No. M21K3c0130).

\appendices

\section{More details of X-DualNet}

\subsection{Network architecture}
The detailed architecture of the generator $G$ in our \textit{X-DualNet} is shown in Table \ref{network}. The main generator and two auxiliary generators share the same architecture and network parameters. $G$ takes the vertex coordinates of the identity and pose meshes as inputs. They are fed into the feature extractor which consists of three pointnet++-like convolution layers. Then we can calculate the correspondence matrix with the extracted features using optimal transport. With the calculated matrix, we warp the pose mesh to get the warped mesh. Finally, we generate the output with several elastic instance normalization residual blocks in the pose transfer module.

\linespread{1.25}
\begin{table*}[h]
\centering
\caption{\textbf{The network architecture of the generator in \textit{X-DualNet}}. N is the vertex number, and D means the feature channel number. We show an example when training on the human data SMPL \cite{loper2015smpl} whose vertex number is 6890. Here, ($1 \times 1$) is the kernel size of the convolution layer.}
\label{network}
\begin{tabular}{c|c|c|c}
\hline
                                         & Module                             & Layers                & Output ($N \times D$)     \\ \hline
\multirow{5}{*}{Correspondence Learning} & \multirow{3}{*}{Feature extractor} & Pointnet++-like conv               & $6890 \times 32$   \\ 
                                         &                                    & Pointnet++-like conv                & $6890 \times 64$   \\ 
                                         &                                    & Pointnet++-like conv                & $6890 \times 128$  \\ \cline{2-4} 
                                         & Optimal transport                  & Correspondence matrix & $6890 \times 6890$ \\ \cline{2-4} 
                                         & Warping                            & Warped pose mesh      & $6890 \times 3$    \\ \hline
Pose Transfer               & Mesh generation             & \begin{tabular}[c]{@{}c@{}}Conv1d ($3 \times 3$)\\ Conv1d ($1 \times 1$)\\ ElaIN Resblock\\ Conv1d ($1 \times 1$)\\ ElaIN Resblock\\ Conv1d ($1 \times 1$)\\ ElaIN Resblock\\ Conv1d ($1 \times 1$)\end{tabular} & \begin{tabular}[c]{@{}c@{}}$6890 \times 1024$\\ $6890 \times 1024$\\ $6890 \times 1024$\\ $6890 \times 512$\\ $6890 \times 512$\\ $6890 \times 256$\\ $6890 \times 256$\\ $6890 \times 3$\end{tabular} \\ \hline
\end{tabular}
\end{table*}

\subsection{Solving OT problem with Sinkhorn algorithm}
In this section, we solve the optimal transport (OT) problem defined in the main paper with Sinkhorn algorithm \cite{sinkhorn1967diagonal}. Following \cite{cuturi2013sinkhorn}, we introduce an entropic regularization term to solve the OT problem efficiently, 

\begin{equation}
\begin{aligned}
    \mathbf{T}_{m} = &\mathop{\arg\min}_{\mathbf{T} \in \mathbb{R}_{+}^{N_{id} \times N_{pose}}} \sum_{i j}\mathbf{Z}(i, j)\mathbf{T}(i, j) \\
    &+ \varepsilon\mathbf{T}(i, j)(\log\mathbf{T}(i, j) - 1)   \\
    s.t. \quad \mathbf{T}\mathbf{1}_{N_{pose}} &= \mathbf{1}_{N_{id}}N^{-1}_{id}, \quad \mathbf{T}^{\top}\mathbf{1}_{N_{id}} = \mathbf{1}_{N_{pose}}N^{-1}_{pose}.
    \label{transport}
\end{aligned}
\end{equation}
where $\mathbf{T}$, $\mathbf{Z}$ and $\mathbf{T}_{m}$ are the transport matrix, cost matrix, and optimal matching matrix respectively, $\mathbf{1}_{N_{id}} \in \mathbb{R}^{N_{id}}$ and $\mathbf{1}_{N_{pose}} \in \mathbb{R}^{N_{pose}}$ are vectors whose elements are all $1$, $\varepsilon$ is the regularization parameter. The details of the solving process are shown in Algorithm \ref{ot}.
\renewcommand{\algorithmicrequire}{\textbf{Input:}}
\renewcommand{\algorithmicensure}{\textbf{Output:}}
\begin{algorithm}
  \caption{Optimal transport problem with Sinkhorn algorithm.}
  \label{ot}
  \begin{algorithmic}
    \REQUIRE Cost matrix $\mathbf{Z}$, regularization parameter $\varepsilon$, iteration number $i_{max}$.          
    \ENSURE Optimal matching matrix $\mathbf{T}_{m}$.    
    \STATE $\mathbf{U} \gets \mbox{exp}(-\mathbf{Z}/\varepsilon)$;
    \STATE $\mathbf{a} \gets \mathbf{1}_{N_{id}}N_{id}^{-1}$;
    \FOR{$i = 0, ..., i_{max}-1$}
    \STATE $\mathbf{b} \gets (\mathbf{1}_{N_{pose}}N_{pose}^{-1})/(\mathbf{U}^{\top} \mathbf{a})$;
    \STATE $\mathbf{a} \gets (\mathbf{1}_{N_{id}}N_{id}^{-1})/(\mathbf{U}\mathbf{b})$;
    
    \ENDFOR 
    \STATE $\mathbf{T}_{m}  \gets \mathrm{diag}(\mathbf{a})\mathbf{U}\mathrm{diag}(\mathbf{b}).$
  \end{algorithmic}
\end{algorithm}

\subsection{More implementation details}
For 3D-CoreNet, $\lambda_{rec}$ for the reconstruction loss is set as 2000. We implement it with Pytorch and use Adam optimizer. 3D-CoreNet is trained for 200 epochs on one RTX 3090 GPU, the learning rate is fixed at $1e\mbox{-}4$ in the first 100 epochs and decays $1e\mbox{-}6$ each epoch after 100 epochs. The batch size is 8. The training time is about 24 hours for the human data and about 36 hours for the animal data.

In X-DualNet, for the as-rigid-as-possible (ARAP) deformer, we choose to fix $10\%$ vertices of the output mesh and deform the identity mesh to get the deformed output. We run ARAP for 50 iterations in the last 50 epochs. We train our model on two RTX 3090 GPUs. The batch size is 4 for the human data and 6 for the animal data. X-DualNet needs about 90 hours for training SMPL and about 100 hours for training the SMAL. In Algorithm \ref{ot}, we set $\varepsilon = 0.03$ and $i_{max} = 5$. $\epsilon$ in Equation 5 in the main paper is $1e\mbox{-}5$. 

\section{Proposed unsupervised baseline}
\begin{figure*}
  \centering
  \includegraphics[scale=0.38]{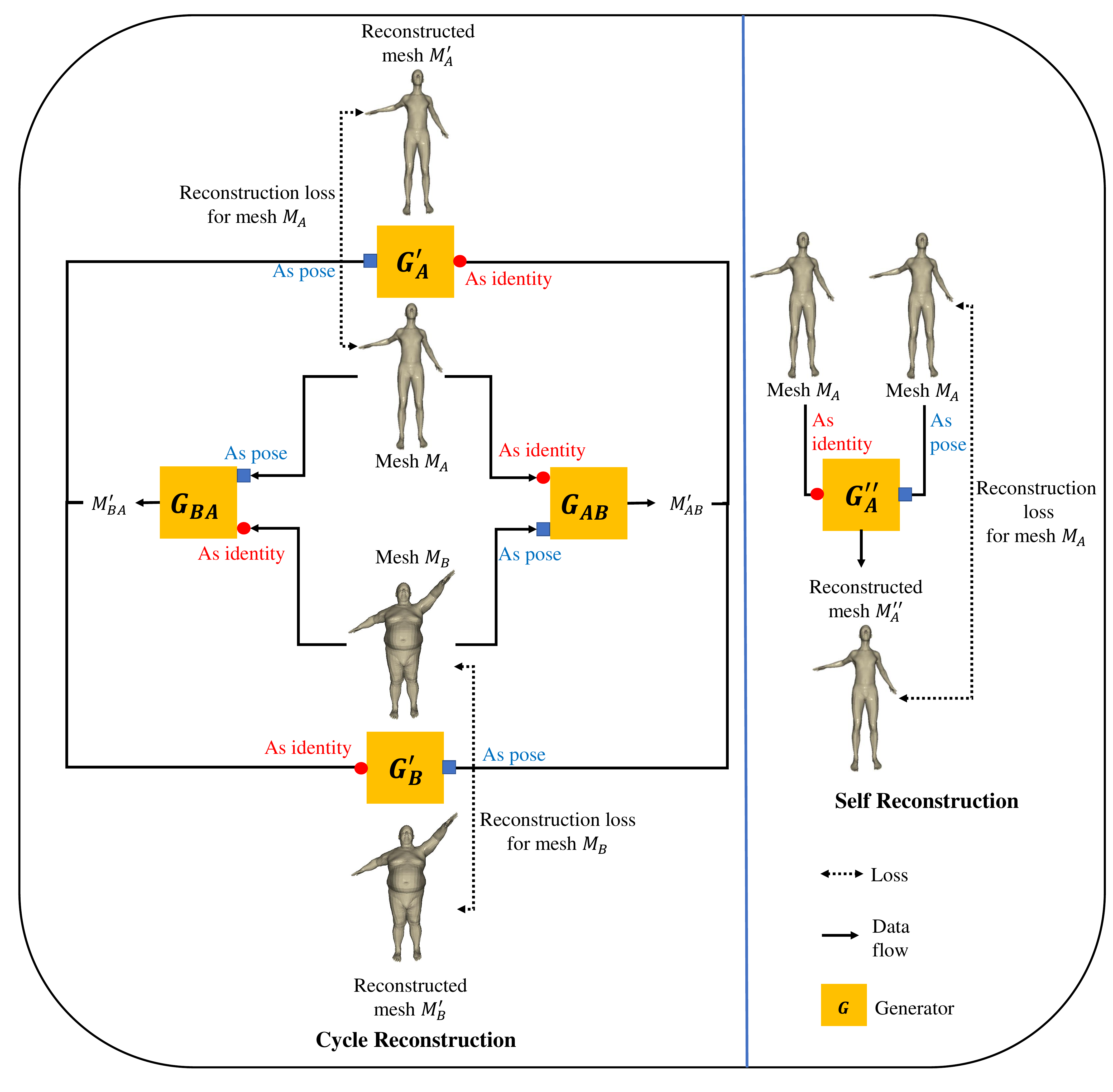}
  \caption{\textbf{The design of our proposed unsupervised baseline}. It includes two modules: cycle reconstruction and self reconstruction. In the cycle reconstruction module, given two meshes $M_{A}$ and $M_{B}$, we generate $M^{\prime}_{AB}$ and $M^{\prime}_{BA}$ with two generators $G_{AB}$ and $G_{BA}$ respectively. Here, $M^{\prime}_{AB} = G_{AB}(M_{A}, M_{B})$ and $M^{\prime}_{BA} = G_{BA}(M_{B}, M_{A})$. Then we reconstruct $M_{A}$ with the generator $G^{\prime}_{A}$. Specifically, $G^{\prime}_{A}$ takes $M^{\prime}_{AB}$ as the identity mesh and $M^{\prime}_{BA}$ as the pose mesh respectively to generate $M^{\prime}_{A} = G^{\prime}_{A}(M^{\prime}_{AB}, M^{\prime}_{BA})$. Similarly, we can reconstruct $M^{\prime}_{B}$ with another generator $G^{\prime}_{B}$. For $G^{\prime}_{B}$, we view $M^{\prime}_{BA}$ as the identity mesh and $M^{\prime}_{AB}$ as the pose mesh respectively to generate $M^{\prime}_{B} = G^{\prime}_{B}(M^{\prime}_{BA}, M^{\prime}_{AB})$. For the self reconstruction module, we take $M_{A}$ as the identity mesh and the pose mesh to reconstruct $M^{\prime\prime}_{A}$. All of the five generators share the same network parameters.}
  \label{uncs}
\end{figure*}

We introduce the network design of the proposed unsupervised baseline. As shown in Figure \ref{uncs}, it includes two modules: cycle reconstruction and self reconstruction. In the cycle reconstruction module, given two meshes $M_{A}$ and $M_{B}$, we generate $M^{\prime}_{AB}$ and $M^{\prime}_{BA}$ with two generators $G_{AB}$ and $G_{BA}$ respectively. Here, $M^{\prime}_{AB} = G_{AB}(M_{A}, M_{B})$ and $M^{\prime}_{BA} = G_{BA}(M_{B}, M_{A})$. Then we reconstruct $M_{A}$ with the generator $G^{\prime}_{A}$. Specifically, $G^{\prime}_{A}$ takes $M^{\prime}_{AB}$ as the identity mesh and $M^{\prime}_{BA}$ as the pose mesh respectively to generate $M^{\prime}_{A} = G^{\prime}_{A}(M^{\prime}_{AB}, M^{\prime}_{BA})$. Similarly, we can reconstruct $M^{\prime}_{B}$ with another generator $G^{\prime}_{B}$. For $G^{\prime}_{B}$, we view $M^{\prime}_{BA}$ as the identity mesh and $M^{\prime}_{AB}$ as the pose mesh respectively to generate $M^{\prime}_{B} = G^{\prime}_{B}(M^{\prime}_{BA}, M^{\prime}_{AB})$. 

For the self reconstruction module, we take $M_{A}$ as the identity mesh and the pose mesh to reconstruct $M^{\prime\prime}_{A}$. Without the self reconstruction module, the generated outputs are always the mirror of desired results or flip the pose mesh at an abnormal angle.

All five generators share the same network parameters. In this proposed unsupervised baseline, we adopt five generators and three reconstruction losses to instruct the training and encourage $M^{\prime}_{A} \approx M_{A}$, $M^{\prime}_{B} \approx M_{B}$ and $M^{\prime\prime}_{A} \approx M_{A}$. However, our \textit{X-DualNet} still has a better performance than it although we do not add the ARAP deformer, as discussed in the experiment section in the main paper. For the proposed baseline, the reconstructed meshes are predicted from generated meshes $M^{\prime}_{BA}$ and $M^{\prime}_{AB}$, which can make the training more difficult. For \textit{X-DualNet}, the reconstructed meshes are predicted from one input mesh and one generated mesh, which provide more known information for training. Therefore, X-DualNet is easier to train and has a better performance than the baseline. And since the baseline contains 5 generators, it needs more time for training.

\section{More experimental results}

\subsection{Generated results on the human and animal data}
\begin{figure*}[h]
  \centering
  \includegraphics[scale=1.1]{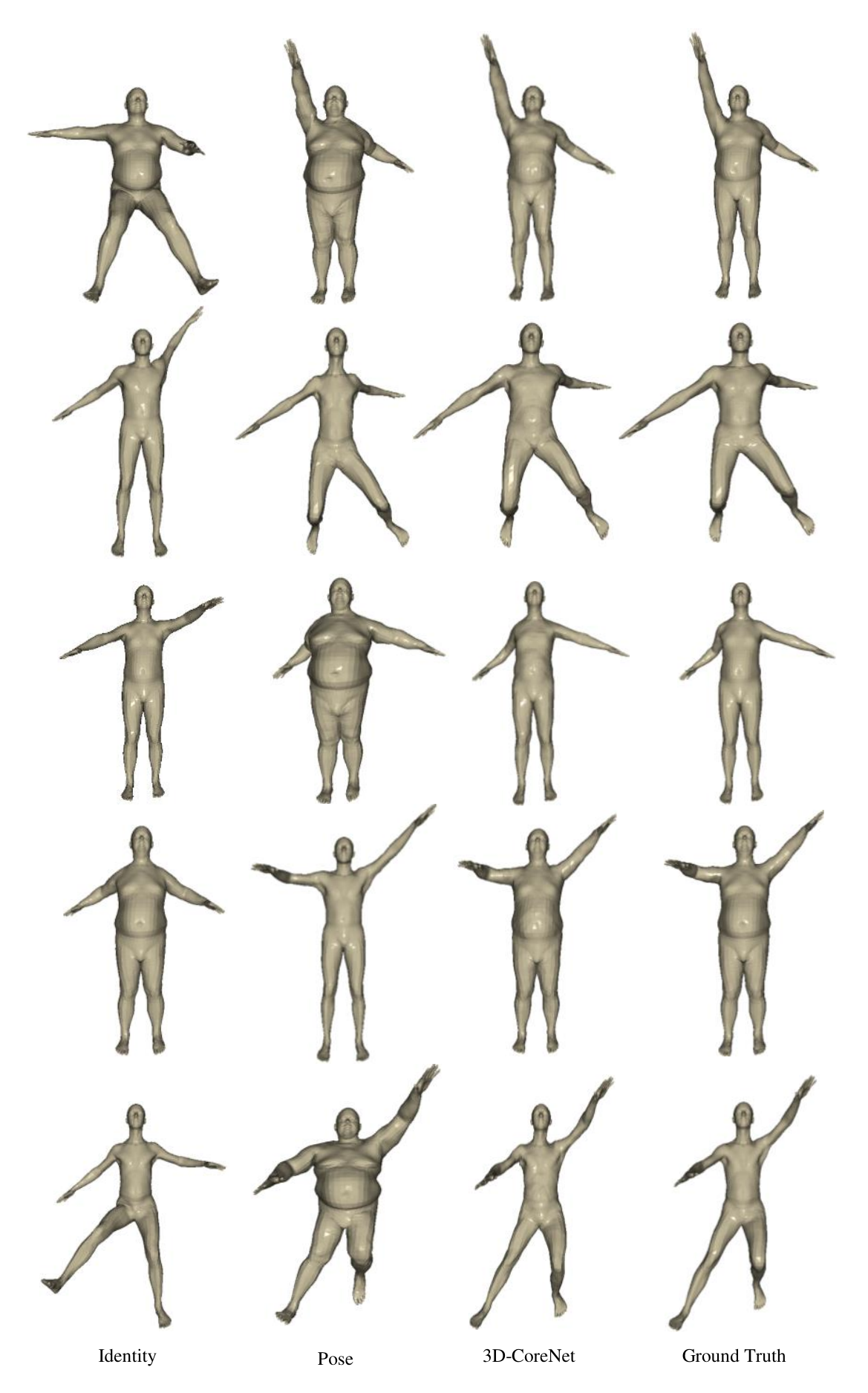}
  \caption{\textbf{More results generated by 3D-CoreNet on the human data SMPL \cite{loper2015smpl}}.}
  \label{moresmpl}
\end{figure*}

\begin{figure*}[h]
  \centering
  \includegraphics[scale=0.35]{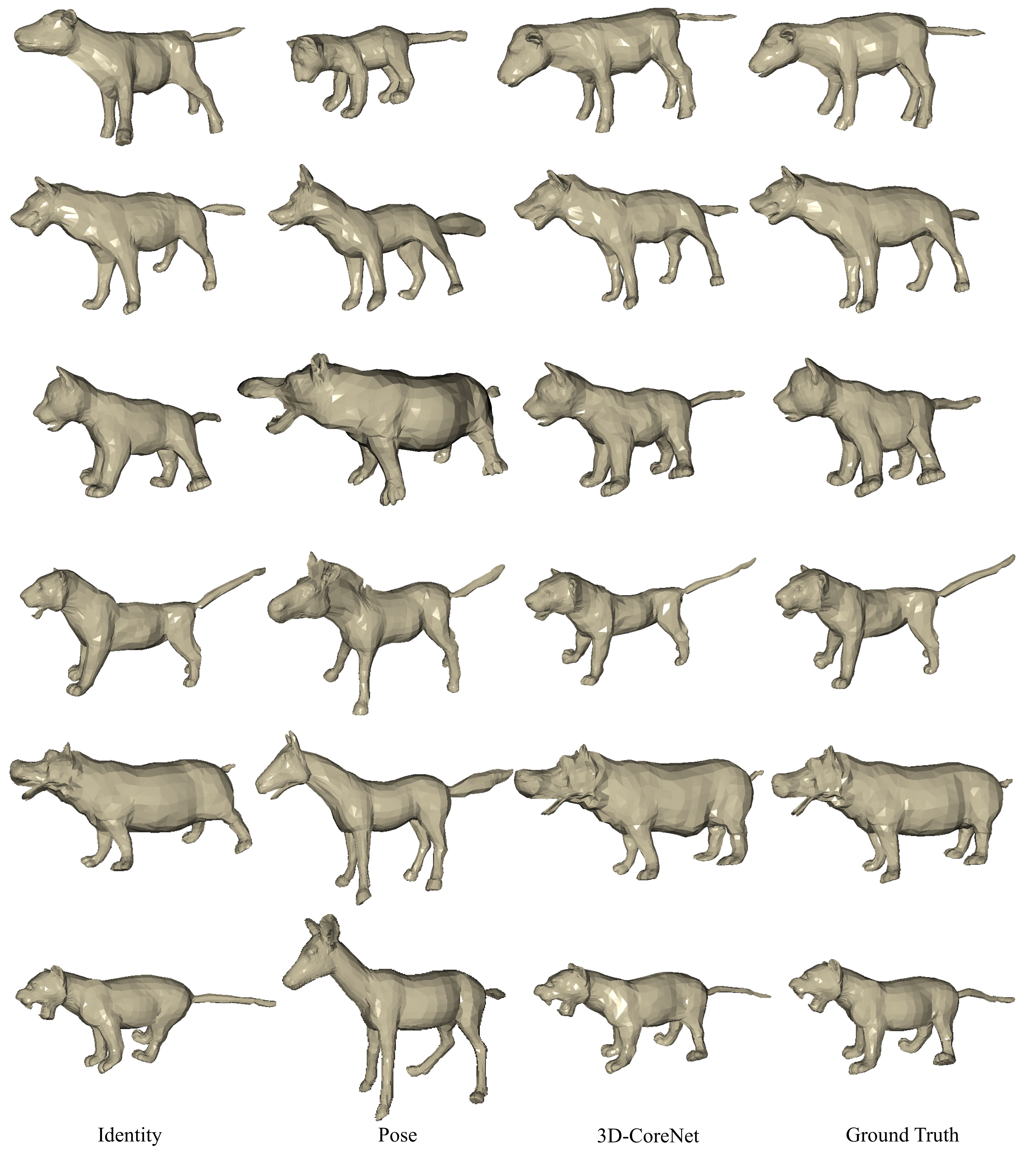}
  \caption{\textbf{More results generated by 3D-CoreNet on the animal data SMAL \cite{zuffi20173d}}.}
  \label{moresmal}
\end{figure*}

\begin{figure*}[h]
  \centering
  \includegraphics[scale=1.1]{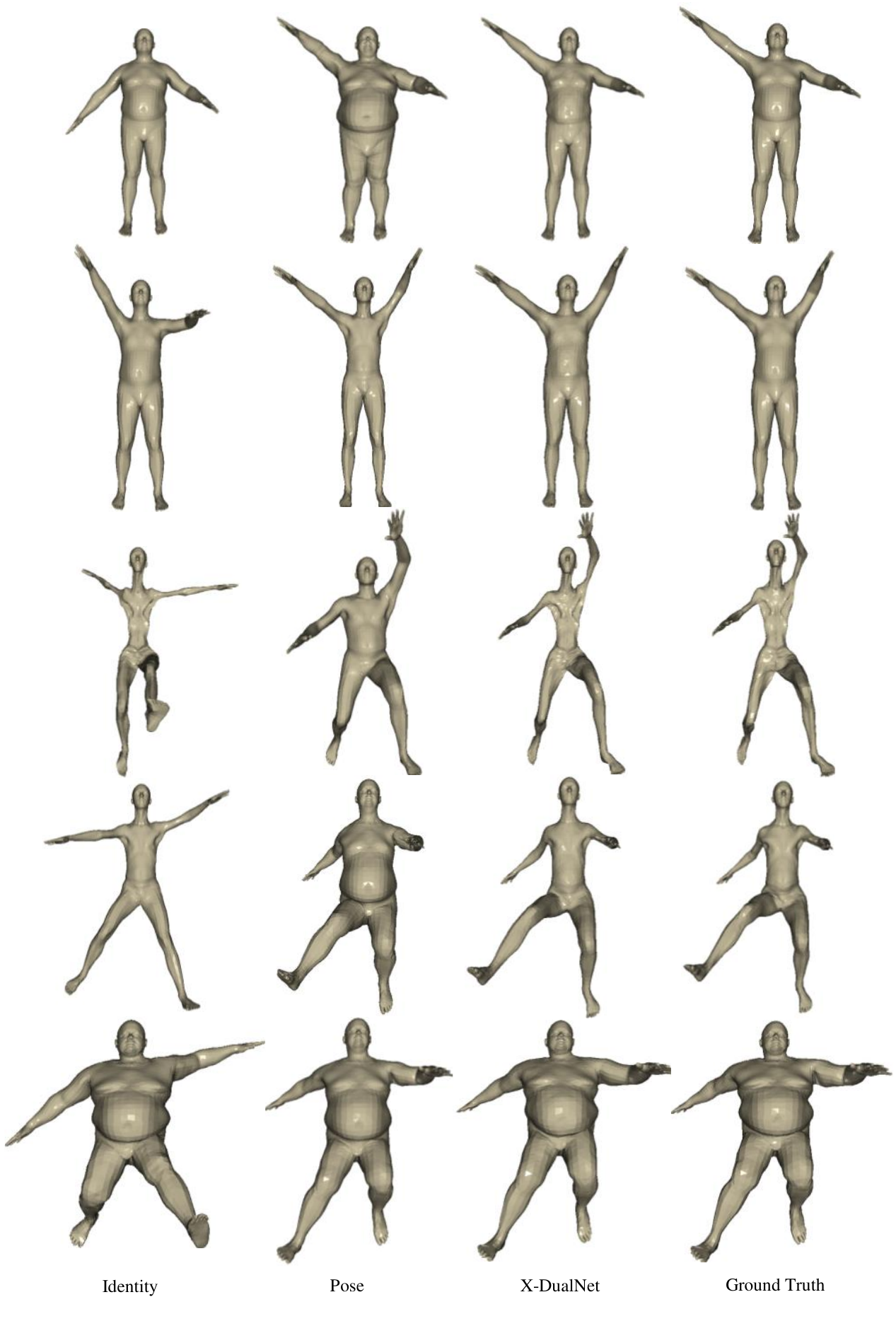}
  \caption{\textbf{More results generated by our \textit{X-DualNet} on the human data SMPL \cite{loper2015smpl}}.}
  \label{moresmpl-un}
\end{figure*}

\begin{figure*}[h]
  \centering
  \includegraphics[scale=0.35]{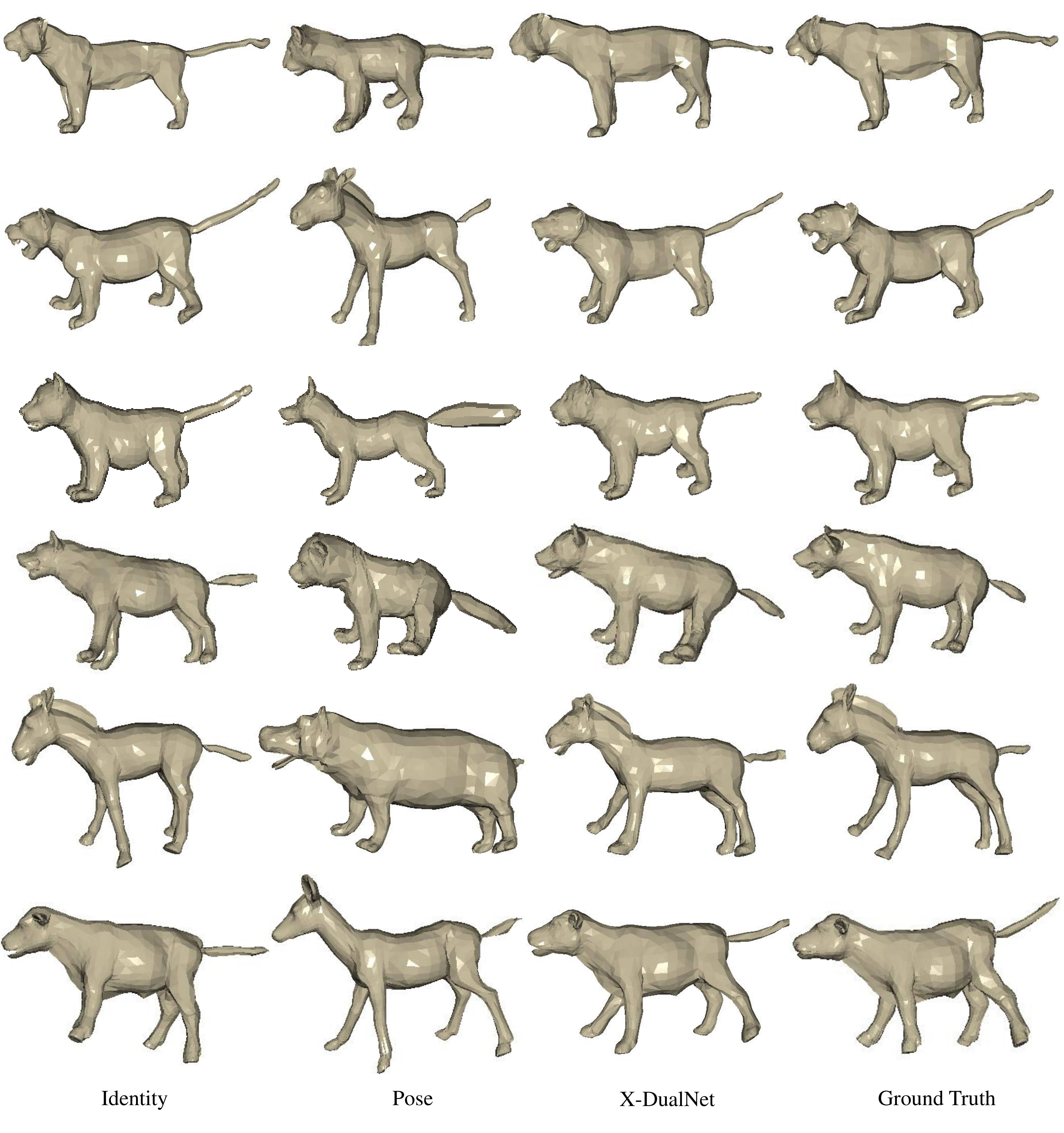}
  \caption{\textbf{More results generated by our \textit{X-DualNet} on the animal data SMAL \cite{zuffi20173d}}.}
  \label{moresmal-un}
\end{figure*}

In this section, we generate more results with 3D-CoreNet and X-DualNet. In Figure \ref{moresmpl} and Figure \ref{moresmal}, we show more results generated by our 3D-CoreNet on the human data SMPL \cite{loper2015smpl} and the animal data SMAL \cite{zuffi20173d} respectively. In Figure \ref{moresmpl-un} and Figure \ref{moresmal-un}, we show more results generated by our X-DualNet on SMPL and SMAL respectively. 

\subsection{Learned shape correspondence}
\begin{figure*}[h]
  \centering
  \includegraphics[scale=0.5]{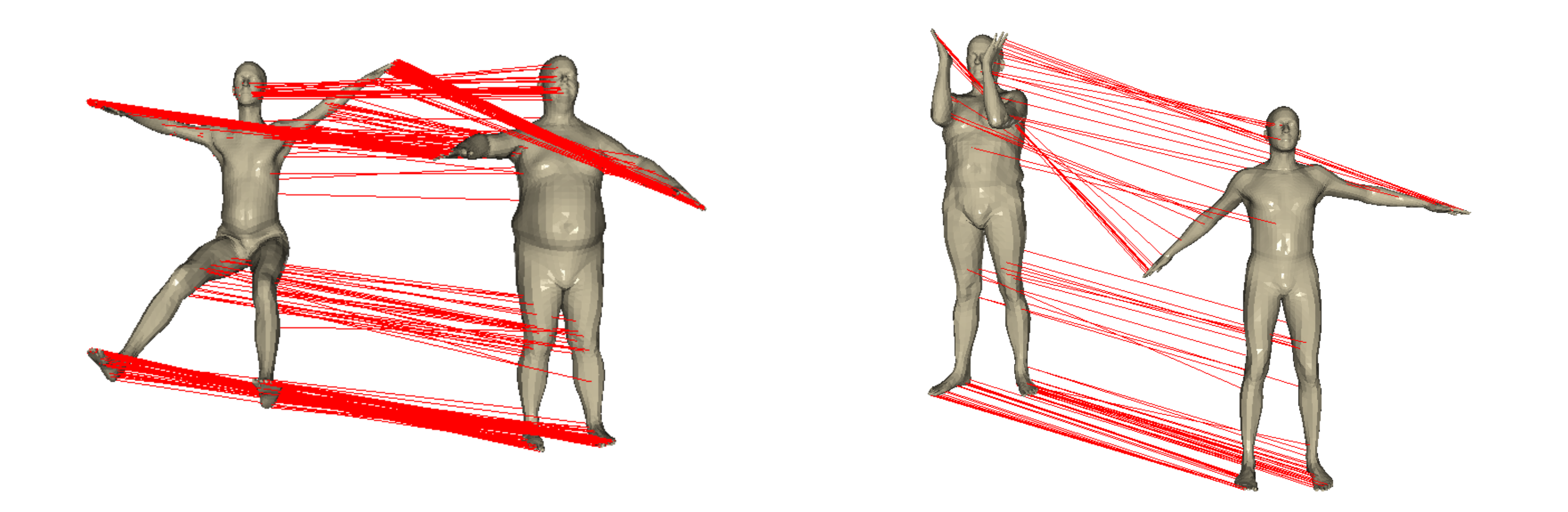}
  \caption{\textbf{The learned shape correspondence between human meshes}. We choose several corresponding vertices on different human meshes for visualization. Note that the vertices of the meshes are shuffled randomly before input. In the first group, the human meshes are from SMPL \cite{loper2015smpl}. In the second group, the left mesh is from FAUST \cite{bogo2014faust}, and the right mesh is from SMPL \cite{loper2015smpl}.}
  \label{humancorr}
\end{figure*}

\begin{figure*}[h]
  \centering
  \includegraphics[scale=0.6]{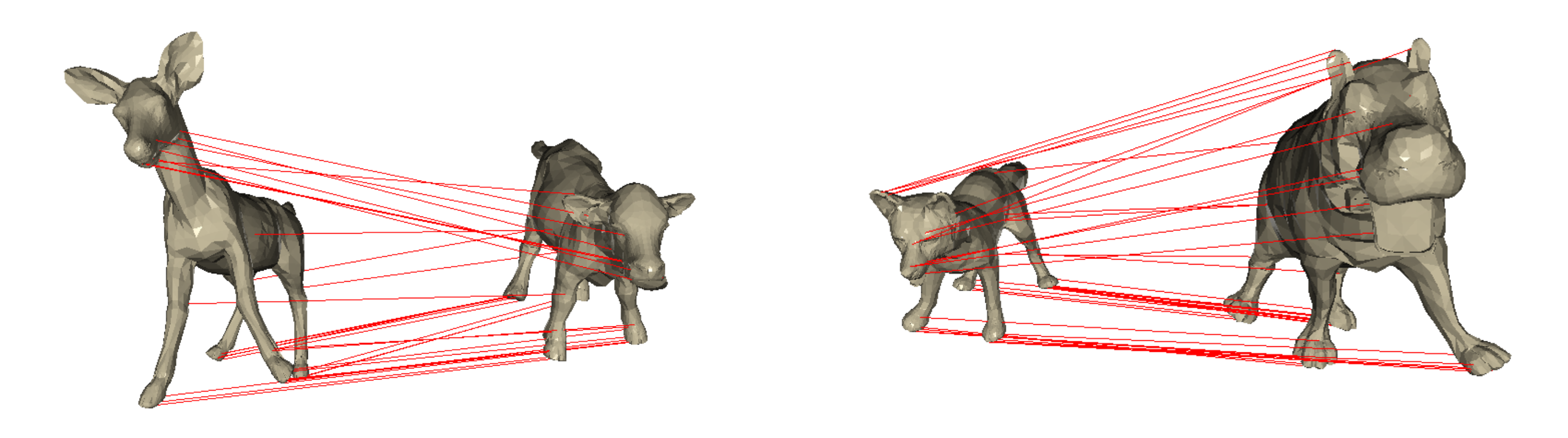}
  \caption{\textbf{The learned shape correspondence between animal meshes}. We choose several corresponding vertices on different animal meshes for visualization. Note that the vertices of the meshes are shuffled randomly before input. The four meshes are all from SMAL \cite{zuffi20173d}.}
  \label{animalcorr}
\end{figure*}

As shown in Figure \ref{humancorr} and Figure \ref{animalcorr}, we visualize the shape correspondence learned by X-DualNet with red lines between different human and animal meshes. Note that the vertices of the meshes are shuffled randomly before input.

\subsection{Average inference times}
In this section, we compare the average inference times for every pose transfer of different methods in the same experimental settings. As shown in Table \ref{time}, the traditional deformation transfer method \cite{sumner2004deformation} takes the longest time compared to the deep learning-based methods. For NPT \cite{wang2020neural}, they do not learn the correspondence between meshes, so they have the shortest inference time but the generation performance is degraded. 3D-CoreNet and X-DualNet achieve notable improvements in generating high-quality results while the inference time is also acceptable. They share the same inference time since their inference only needs to go through one single generator.

\begin{table}[h]
\centering
\caption{\textbf{Average inference times of different methods}.} 
\label{time}
\begin{tabular}{cccc}
\hline
Method & DT  & NPT & 3D-CoreNet \& X-DualNet \\ \hline
Time   &    3.3352s      &      0.0068s    &     0.1154s                                       \\ \hline
\end{tabular}
\end{table}

\subsection{Generalization capability}
\begin{figure*}[h]
  \centering
  \includegraphics[scale=0.26]{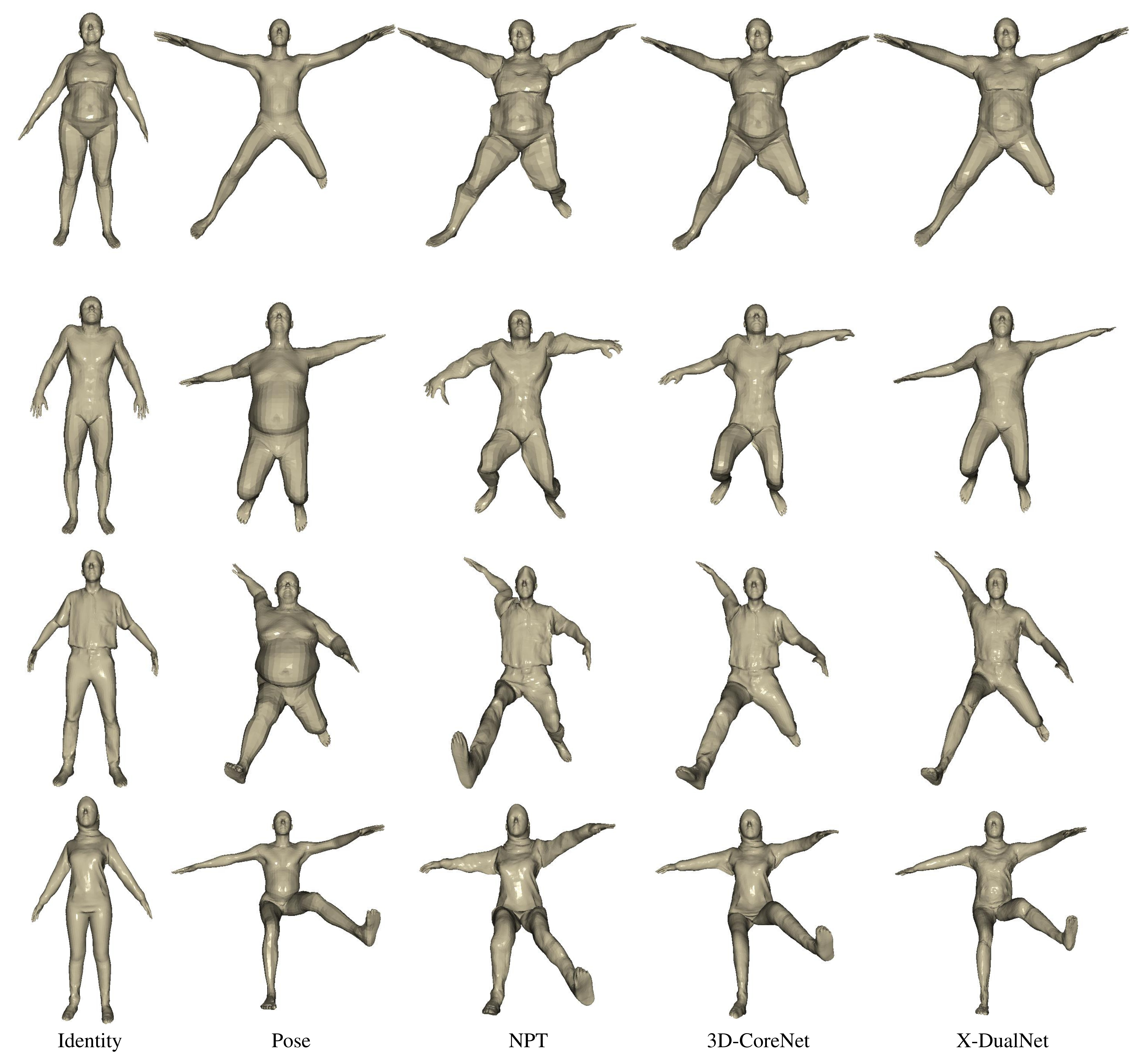}
  \caption{\textbf{More results on FAUST \cite{bogo2014faust} and MGN \cite{bhatnagar2019multi}}. The identity meshes are from FAUST and MGN. We compare X-DualNet with NPT \cite{wang2020neural} and 3D-CoreNet to test the generalization capability.}
  \label{faustmg}
\end{figure*}

In Figure \ref{faustmg}, we show more results generated by X-DualNet on FAUST \cite{bogo2014faust} and MGN \cite{bhatnagar2019multi}. To further test the generalization capability of the model, we compare X-DualNet with NPT \cite{wang2020neural} and 3D-CoreNet. As we can see, the results generated by X-DualNet are more smooth and realistic. The results generated by NPT and 3D-CoreNet always have some artifacts.   

\subsection{Robustness to noise}
\begin{figure*}[h]
  \centering
  \includegraphics[scale=0.433]{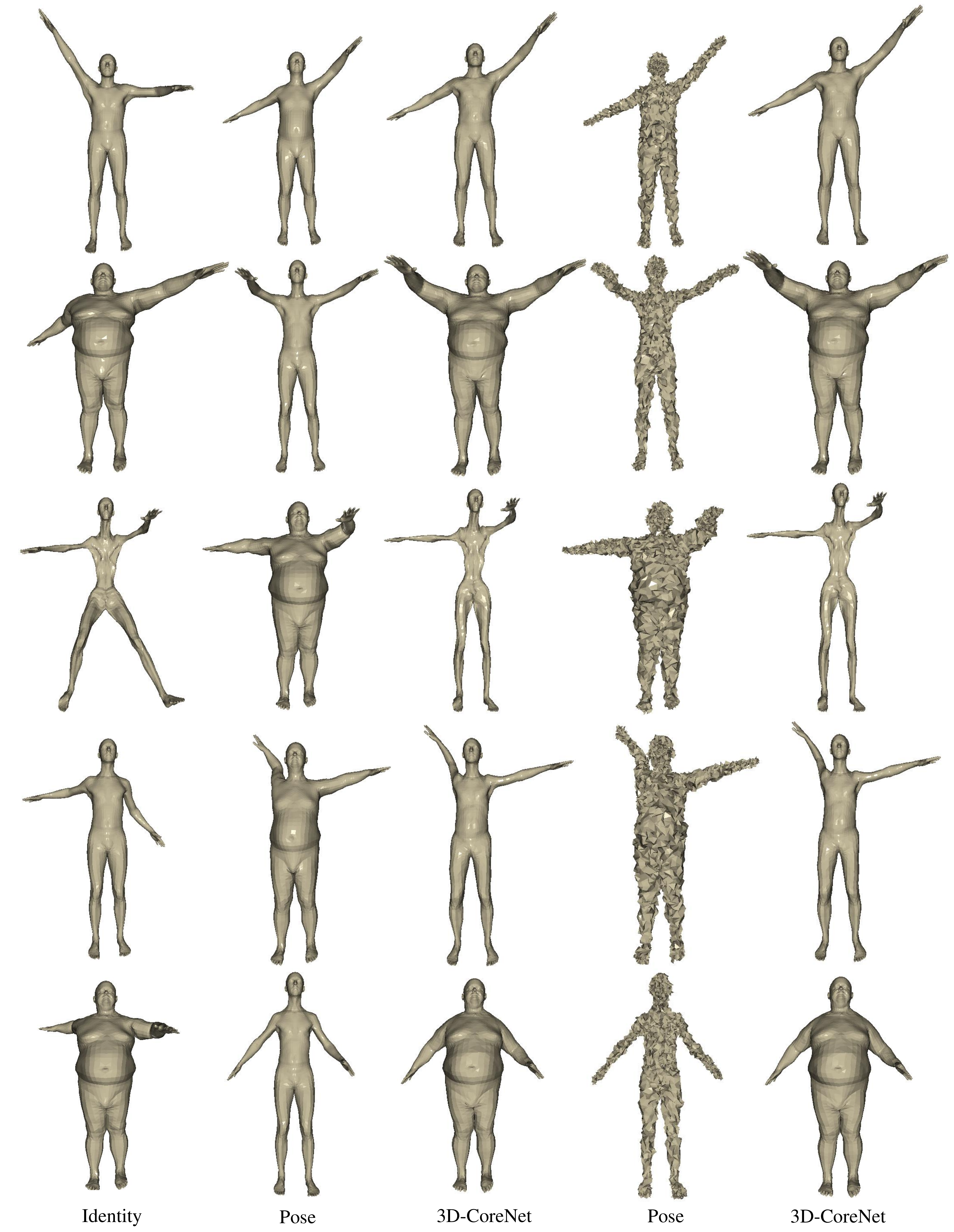}
  \caption{\textbf{Robustness to noise}. Here, we add noise to the pose mesh. The model can still produce high-quality results.}
  \label{noise}
\end{figure*}

To test the robustness of our model, we add noise to the vertex ordinates of the pose mesh. The results generated by 3D-CoreNet are shown in Figure \ref{noise}, the model still produces high-quality results.

\subsection{Limitations}
Although our method produces satisfactory results in most cases and has better performance than previous works, there are still some limitations that need to be solved in the future. We will discuss some failure cases in the following.

When testing our model on the animal data as shown in Figure \ref{moresmal-un}, the tails of the animals are difficult to preserve (the fifth row and the sixth row). When evaluating \textit{X-DualNet} on two meshes that one is wearing clothes and the other is not, such as the third row in Figure \ref{faustmg}, X-DualNet does not keep the clothes very well although it transfers the pose successfully compared to 3D-CoreNet.

\ifCLASSOPTIONcaptionsoff
  \newpage
\fi



\bibliographystyle{IEEEtran}
%
\bibliography{egbib}

\end{document}